\documentclass[journal]{IEEEtran}
\usepackage{cite}

\usepackage{color}

\usepackage{amsmath}
\usepackage{algorithm}
\usepackage{algorithmic}
\usepackage{bm}
\usepackage{latexsym}
\usepackage{amsthm}
\usepackage{url}
\usepackage{amsfonts}
\usepackage{amssymb}

\usepackage{indentfirst}
\usepackage{float}
\usepackage{array}
\usepackage{multirow}
\usepackage{graphicx}  
\usepackage{subfig}
\usepackage{booktabs}
\usepackage{multirow}
\usepackage{rotating}

\makeatletter \@fpsep = 1pt \makeatother

\begin{document}

\title{Adaptive Locality Preserving Regression}
%
%
%

\author{Jie Wen,
        Zuofeng Zhong,
        Zheng Zhang,
        Lunke Fei$^*$,
        Zhihui Lai,
        Runze Chen

\thanks{This work was supported in part by the National Natural Science Foundation of China under Grant nos. 61702110, 61702117 and 61703169, and in part by Technology Program of Guangzhou under Grant no. 201804010355. (Jie Wen and Zuofeng Zhong are co-first authors with equal contributions.) (Corresponding author: Lunke Fei.)}
\thanks{Jie Wen and Lunke Fei are with the School of Computer Science and Technology, Guangdong University of Technology, Guangzhou 510006, Guangdong, China. (Email: jiewen\_pr@126.com; flksxm@126.com)}
\thanks{Zuofeng Zhong is with the College of Computer Science and Software Engineering, Shenzhen University, Shenzhen 518055, Guangdong, China, and is also with the Institute of Textiles and Clothing, The Hong Kong Polytechnic University, Hong Kong. (Email: zfzhong2010@gmail.com)}
\thanks{Zheng Zhang is with the School of Information Technology \& Electrical Engineering, The University of Queensland, Brisbane, QLD 4072, Australia. (Email: darrenzz219@gmail.com)}
\thanks{Zhihui Lai is with the College of Computer Science and Software Engineering, Shenzhen University, Shenzhen 518055, Guangdong, China. (Email: lai\_zhi\_hui@163.com)}
\thanks{Runze Chen is with the School of Computer Science and Technology, Harbin Institute of Technology, Shenzhen, Shenzhen 518055, China. (Email: chenxiaoyue98@gmail.com)}
}

\maketitle

\begin{abstract}
This paper proposes a novel discriminative regression method, called adaptive locality preserving regression (ALPR) for classification. In particular, ALPR aims to learn a more flexible and discriminative projection that not only preserves the intrinsic structure of data, but also possesses the properties of feature selection and interpretability. To this end, we introduce a target learning technique to adaptively learn a more discriminative and flexible target matrix rather than the pre-defined strict zero-one label matrix for regression. Then a locality preserving constraint regularized by the adaptive learned weights is further introduced to guide the projection learning, which is beneficial to learn a more discriminative projection and avoid overfitting. Moreover, we replace the conventional `Frobenius norm' with the special $l_{2,1}$ norm to constrain the projection, which enables the method to adaptively select the most important features from the original high-dimensional data for feature extraction. In this way, the negative influence of the redundant features and noises residing in the original data can be greatly eliminated. Besides, the proposed method has good interpretability for features owning to the row-sparsity property of the $l_{2,1}$ norm. Extensive experiments conducted on the synthetic database with manifold structure and many real-world databases prove the effectiveness of the proposed method.
\end{abstract}


\IEEEpeerreviewmaketitle
\section{Introduction}
\IEEEPARstart{R}{egression} analysis focuses on estimating the relationships among the dependent variables and independent variables, which has aroused much attention in fields of machine learning \cite{lai2018generalized,qi2017joint,jiang2018ensemble,qi2011locality,kang2018systematic,liu2017deep,li2017shared,deng2017novel}. For supervised classification, one of the major tasks is to learn a proper mapping that precisely transforms the training data into their labels. To this end, various regression analysis methods have been proposed over the past decades, such as the ridge regression \cite{hoerl1970ridge,dhillon2013risk}, partial least squares \cite{krishnan2011partial}, modified minimum squared error \cite{xu2014modified}, and least square regularized regression \cite{xu2013least,zeng2018robust}, etc. Besides, many kernel based regression methods, such as kernel ridge regression \cite{an2007face} and support vector regression \cite{smola2004tutorial}, have also been proposed for the non-separable cases.

Most of the conventional methods prefer to exploit the pre-defined zero-one label matrix as the regression target. However, this simple label matrix is not the optimal discriminative target for supervised classification \cite{xiang2012discriminative,wen2018inter,yi2017unified,sun2018heat,xue2013foreground,zhang2017marginal}. First, it limits the flexibility of projection learning because it is too strict. Second, it cannot push samples of different classes far away because distances of the correct and incorrect label vectors are constant (\emph{i.e.}, $\sqrt 2$) in the target space. To solve the problem, many researchers proposed to learn a more discriminative regression target rather than used the strict zero-one label matrix for regression analysis. For example, in \cite{xiang2012discriminative}, the $\varepsilon$-dragging technique is introduced to adaptively enlarge the distance between the correct and incorrect classes. In \cite{wen2018inter}, a sparse error term is also introduced to relax the strict label matrix, which in turn improves the flexibility of regression analysis. Zhang et al. proposed a very novel target learning technique, in which the target matrix is adaptively learned with large margins between different classes during the regression and projection learning \cite{zhang2015retargeted}.

In most cases, these improved regression based methods can learn a more discriminative projection and obtain a better performance. However, for data with manifold structure or noises, these methods will fail. This is mainly because that (1) these methods only focus on minimizing the regression errors while ignoring the intrinsic geometric structure of data, which may destroy the structure of the original data and lead to overfitting; (2) All features including the important features and noises are treated equally in these methods, which cannot guarantee these methods to obtain a clean projection.

In recent years, many researchers have also discovered the first issue and have made many efforts to address it. For instance, the low-rank linear regression (LRLR) tries to uncover the low-rank structure hidden in the high-dimensional data for regression \cite{cai2013equivalent}. In \cite{wen2018inter}, a novel inter-class sparse constraint is introduced, which tries to guarantee the common structure with respect to each class. Besides, integrating the manifold learning into the regression framework is a well-received approach to avoid overfitting \cite{xue2009discriminatively,fang2017regularized}. Xue et al. constructed two graphs according to the label information and local nearest neighbor information to guide the projection learning in regression analysis \cite{xue2009discriminatively}. In \cite{fang2017regularized}, a strict label based graph is introduced to regularize the projection, which allows samples of the same class to be pulled together. Compared with the first two mentioned methods, those manifold learning based methods can preserve the local structure of data better. However, these manifold based methods still have the following shortcomings: (1) all graphs exploited to guide the projection learning are constructed independently with the regression in advance, which cannot guarantee the global optimal projection. (2) These methods are sensitive to noise. When data contain noises, the constructed graph will be incorrect. In this case, it is obviously impossible to learn a discriminative projection with the guiding of the incorrect graph. (3) These methods cannot preserve the same nearest neighbor ranks as the original data since they exploit the same weight, \emph{i.e.}, 1, to regularize all nearest neighbors while ignoring the differences of similarity degree among these nearest neighbor pairs. In other words, they cannot preserve the intrinsic nearest neighbor structure of data.

In this paper, we propose a novel and effective method to solve the above problems and learn a more discriminative projection for classification. Specially, the proposed method introduces a novel graph regularization term into the regression framework, in which the graph is adaptively learned in a supervised style and then in turn guides the projection learning. In this way, the proposed method has the potential to learn the global optimal projection that not only can fit the label well, but also can preserve the intrinsic nearest neighbor structure of each class. To improve the flexibility of projection learning, the retargeted learning technique is introduced to our model. Most importantly, a row-sparsity norm instead of the conventional `Frobenius' norm is introduced to constrain the projection, which enables the method to reduce the negative influence of the redundant features and noises. By artfully integrating the above terms into one regression framework, the proposed method is encouraged to perform better. In summary, our work has the following contributions:

(1) We propose a novel regression framework that integrates the adaptive locality preserving, feature selection, and discriminative target learning. Compared with the other methods, the proposed method can learn a more reasonable and discriminative projection and avoid the overfitting problem.

(2) We introduce a novel supervised graph learning and embedding constraint, which can discover the intrinsic local geometric structures of data and adaptively learn the similarity weights to nearest neighbor pairs. By exploiting the reliable local geometric information to regularize the projection, the proposed method has the potential to preserve the intrinsic nearest neighbor structure of data.

(3) The proposed method can adaptively select the most discriminative features for regression by introducing a row-sparsity constraint. This allows the method to reduce the negative influence of noise and improves the interpretability of projection.

This paper is an extended work of our conference paper \cite{Wen2018Adaptive}. Compared with our previous version in \cite{Wen2018Adaptive}, (1) we add more experiments and analyses to prove its effectiveness and convergence; (2) We give deep analyses to the computational complexity, convergence, and demonstrate its superior properties through theoretically comparing some related methods; (3) We analyze the parameter selection in detail; (4) Some Theorems and propositions are provided for readers to better understand our paper; (5) A figure has also been added to show our method.

The paper is organized as follows: In Section II, some notations and several related works are briefly described. In Section III, we mainly present the proposed method and its optimization processes. Section IV analyzes the proposed method in depth. Section V conducts several experiments to prove the effectiveness of the proposed method. Section VI offers the conclusion of the paper.
\section{Related work}
\subsection{Notations}
For convenience, some notations used through the paper are briefly described in this section. In our paper, matrix and vector are denoted by the uppercase letter (e.g. $X$) and lowercase letter (e.g. $x$), respectively. For a matrix $X$, we use $X_{i,j}$ to denote its $i$th row and $j$th column element, and use $X_{i.:}$ and $X_{:,j}$ to represent its $i$th row vector and $j$th column vector, respectively. Some typical norms of matrix $X \in R^{m \times n}$, such as the `Frobenius norm' (\emph{i.e.}, $||X||_{F}$), nuclear norm (\emph{i.e.}, $||X||_*$), $l_1$ norm, and $l_{2,1}$ norm, are defined as: ${\left\| X \right\|_F} = \sqrt {\sum\limits_{i = 1}^m {\sum\limits_{j = 1}^n {X_{i,j}^2} } }$, ${\left\| X \right\|_*} = \sum\nolimits_i {\left| {{\delta _i}} \right|}$, ${\left\| X \right\|_1} = \sum\limits_{i = 1}^m {\sum\limits_{j = 1}^n {\left| {{X_{i,j}}} \right|} }$, and ${\left\| X \right\|_{2,1}} = \sum\limits_{i = 1}^m {\sqrt {\sum\limits_{j = 1}^n {X_{i,j}^2} } }$, respectively, where $\delta _i$ is the $i$th singular value of matrix $X$ \cite{luo2015nuclear,luo2018multi,deng2017study,lu2016low}. For a vector $x$ with $m$ elements, its $l_2$ norm is defined as ${\left\| x \right\|_2} = \sqrt {\sum\limits_{i = 1}^m {{(x_i)}^2} } $, where $x_i$ denotes its $i$th element. The trace operation of matrix is denoted by $Tr(\cdot)$. $I$ denotes the identity matrix. $\textbf{\emph{1}}$ is a column vector, where all elements are 1. $X^{-1}$ and $X^T$ are the inverse matrix and transposed matrix of $X$ \cite{sun2017accelerated}, respectively.
\subsection{Linear regression and retargeted least square regression}
Linear regression (LR) is one of the most popular supervised classification methods in fields of machine learning. The objective function of LR is generally formulated as follows \cite{zhang2017discriminative,wang2016msdlsr}:
\begin{equation}\label{1}
\mathop {\min }\limits_W \left\| {Y - X^TW} \right\|_F^2 + {\lambda}\left\| W \right\|_F^2
\end{equation}
where matrix $X \in R^{m \times n}$ denotes the training set, where each column vector represents a sample, $m$ and $n$ denote the feature dimension and number of training samples, respectively. $Y \in R^{n \times C}$ is the label matrix, in which the $i$th row vector represents the label of the $i$th sample in the training set $X$, $C$ is the class number of the training set. In the conventional LR, label matrix $Y$ is generally defined as a special zero-one matrix according to the class information of samples as follows: if the $i$th sample comes from the $j$th class, then only $Y_{i,j}=1$, and all the other elements $Y_{i,k}=0, k \neq j$. $\lambda$ is a penalty parameter. $W$ is the transformation (or projection) for label prediction. When the projection $W$ is obtained by solving (1), the class label of any test sample $y \in {R^{m \times 1}}$ can be predicted via $k = \mathop {\arg \max }\limits_i {\left( {{y^{T}W}} \right)_i}$, where ${\left( {{y^{T}W}} \right)_i}$ is the $i$th element of vector $(y^{T}W)$.

In \cite{zhang2015retargeted}, Zhang et al. pointed out that using the strict zero-one label matrix as the regression target is harmful to classification and limits the flexibility in the discriminative projection learning. To address this issue, Zhang et al. proposed the retargeted least square regression (ReLSR), which seeks to jointly learn a more discriminative and flexible regression target and projection in one framework as follows:
\begin{equation}\label{2}
\mathop {\min }\limits_{W,T} \left\| {T - {W^T}X} \right\|_F^2 + \lambda \left\| W \right\|_F^2  {\kern 2pt} s.t.{\kern 2pt} {T_{i,{l_i}}} - \mathop {\max }\limits_{j \ne {l_i}} {T_{i,j}} \ge 1
\end{equation}
where $l_i$ is the true class index of the $i$th sample.

We can find that in ReLSR, the margins of the correct and incorrect classes are all larger than 1, which encourages it to increase the separability of data in the target space. Moreover, the adaptively learned target matrix provides more flexibility to learn the discriminative projection.

\subsection{Discriminatively regularized least-squares}
To pull samples of the same class closer and push samples of different classes far away as much as possible, Xue et al. \cite{xue2009discriminatively} proposed a graph regularized method, named discriminatively regularized least-squares (DRLS). DRLS explores the underlying geometric knowledge of the original data to guide the projection learning of linear regression, in which two graphs, \emph{i.e.}, intra-class graph $W^w$ and inter-class graph $W^b$, are pre-constructed and regularized on the projection. The objective function of DRLS is formulated as follows:
\begin{equation}\label{5}
\small
\mathop {\min }\limits_W \frac{1}{2}\left\| {Y - X^TW} \right\|_F^2 + \frac{1}{2}W^TX\left( {\eta {L_w} - \left( {1 - \eta } \right){L_b}} \right){X^T}W
\end{equation}
where $\eta$ ($\eta \in [0,1]$) is a penalty parameter, $L_w$ and $L_b$ are the Laplacian matrices of graph $W^w$ and $W^b$, respectively. For a non-negative graph $S$, its Laplacian matrix is defined as $L = D-\frac{S+S^T}{2}$, where matrix $D$ is a diagonal matrix with the $i$th diagonal element as ${D_{i,i}} = \sum\nolimits_{j = 1} {{\frac{S_{i,j}+S_{j,i}}{2}}}$ \cite{xue2009discriminatively}. In DLSR, the intra-class graph $W^w$ and inter-class graph $W^b$ are respectively constructed as follows according to the nearest neighbor information and label information of data:
\begin{equation}\label{3}
\small
W_{i,j}^w = \left\{ {\begin{array}{*{20}{c}}
{1,}&{if{\kern 1pt} {\kern 1pt} {X_{:,j}} \in {N_w}\left( {{X_{:,i}}} \right) {\kern 1pt} {\kern 1pt} or {\kern 1pt} {\kern 1pt} {X_{:,i}} \in {N_w}\left( {{X_{:,j}}} \right){\kern 1pt} {\kern 1pt} }\\
{0,}&{otherwise}
\end{array}} \right.
\end{equation}
\begin{equation}\label{4}
\small
W_{i,j}^b = \left\{ {\begin{array}{*{20}{c}}
{1,}&{if {\kern 1pt} {\kern 1pt} {X_{:,j}} \in {N_b}\left( {{X_{:,i}}} \right){\kern 1pt} {\kern 1pt} or {\kern 1pt} {\kern 1pt} {X_{:,i}} \in {N_b}\left( {{X_{:,j}}} \right){\kern 1pt} {\kern 1pt}}\\
{0,}&{otherwise}
\end{array}} \right.
\end{equation}
where ${N_w}\left( {{X_{:,i}}} \right)$ denotes the sample set which is composed of the nearest neighbors with the same class to sample $X_{:,i}$, ${N_b}\left( {{X_{:,i}}} \right)$ denotes the sample set which is composed of the nearest neighbors with different classes to sample $X_{:,i}$.
\section{The proposed method}
As we all know, the original data contains much useful information, such as the given label information and the underlying geometric information residing in the data \cite{edraki2018generalized,fu2018research,wenjie2018adaptive,qi2018global}. If we can appropriately utilize these information to guide the projection learning, then a better classification performance will be obtained. This illustrates that how to well discover and explore these information is crucial to learn a more compact and discriminative projection for accurately separating these samples. Based on this motivation, we propose a simple but effective method, named adaptive locality preserving regression (ALPR), to learn a more discriminative and compact projection for classification. Fig.1 shows the flowchart of the proposed method.
\begin{figure*}[htbp]
\centering
\includegraphics [width=4.2in,height = 3.7in]{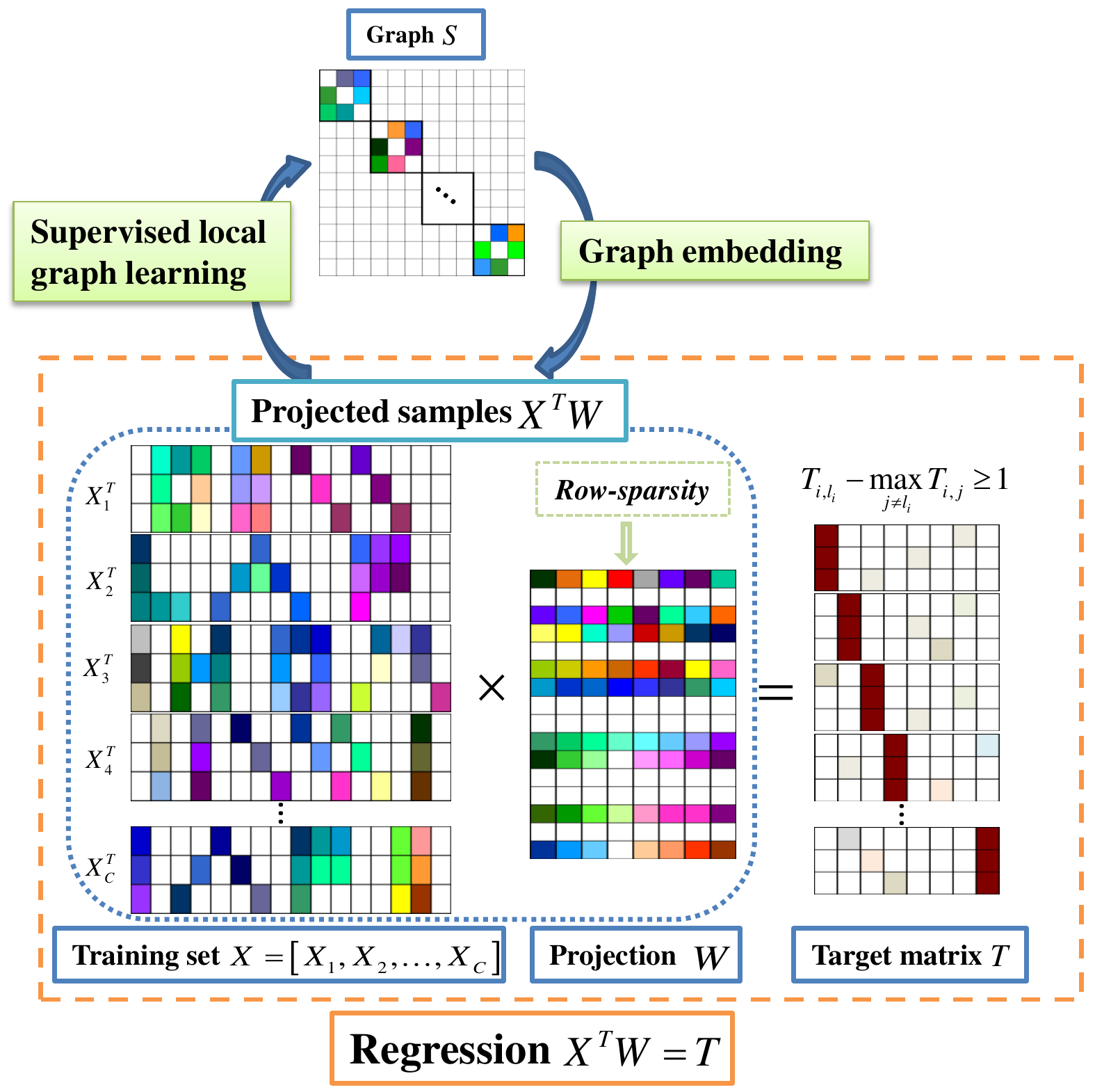}
\caption{The framework of ALPR. ALPR improves the discriminability of the projection by simultaneously considering the following three crucial techniques: 1) introducing the retargeted learning technique to improve the flexibility of regression; 2) imposing the row-sparse constraint on the projection to select the most important features; 3) introducing a novel supervised graph learning and embedding technique to preserve the intrinsic local structure of data.}
\label{fig:arch}
\end{figure*}
\subsection{Model of the proposed method}
Generally, if two samples are nearest neighbors in the original data space, they will be more possible from the same class. Naturally, we also expect that these nearest neighbor relationships can be still preserved in the subspace \cite{wen2018TCYB}. To this end, many methods have been proposed, such as the locality preserving projections \cite{he2004locality} and anchorgraph-based locality preserving projection \cite{jiang2016dimensionality}, etc. Besides, in the branch of LR, many improved methods have also been proposed by introducing the nearest neighbor information \cite{xue2009discriminatively,fang2017regularized}. For instance, DRLS expect the nearest neighbor samples with the same class label to be pulled closer by embedding two graphs \cite{xue2009discriminatively}. Large amount experiments proved that introducing the nearest neighbor information has the potential to further improve the classification performance. However, there are two issues in these conventional LR methods. First, they generally utilize the pre-constructed graph to guide the projection learning, which cannot guarantee the global optimal projection for regression. Second, they ignore the differences of similar degree among these nearest neighbor samples with the same class label since all nearest neighbor pairs are regularized with the same weight (e.g. 1). This will destroy the original nearest neighbor order in each local space. For example, suppose $x_2$ and $x_3$ are the $1$th and $2$nd nearest neighbors of sample $x_1$, respectively. For the conventional graph regularized methods, their transformed samples, \emph{i.e.}, $W^Tx_2$ and $W^Tx_3$ may not still be the $1$th and $2$nd nearest neighbors of sample $W^Tx_1$ in the discriminant subspace since their regularized weights are the same. Therefore, using the fixed weight to regularize all nearest neighbor pairs is not appropriate. A more reasonable approach is to assign different weights for different nearest neighbor pairs according to their similar degrees. In other words, it is better to give a relatively larger weight to the most nearest neighbor pairs. To this end, we introduce the following graph regularization term to explore the geometric information of data:
\begin{equation}\label{5}
\begin{split}
\mathop {\min }\limits_{W,S} {\lambda _1}\sum\limits_{i = 1}^C {{n_i}{\sum\limits_{j,k = 1,k \neq j}^{{n_i}} {{S{{_{j,k}^i}^2}}\|W^Tx_j^i - W^Tx_k^i\|_2^2} } } {\kern 2pt}\\
 + \|Y - X^TW\|_F^2  {\kern 2pt} s.t. {\kern 2pt}  \sum\limits_{k = 1,k \neq j}^{{n_i}} {S_{j,k}^i}  = 1,S_{j,k}^i \ge 0,
\end{split}
\end{equation}
where $X \in R^{m \times n}$ and $Y \in R^{n \times C}$ are the given training set and corresponding label matrix, $C$ is the class number, $n_i$ is the sample number of the $i$th class. $x_j^i$ and $x_k^i$ denote the $j$th and $k$th sample of the $i$th class, respectively. Accordingly, $S_{j,k}^i$ is the weight to regularize the distance of the corresponding samples $x_j^i$ and $x_k^i$ in the discriminant subspace. $\lambda _1$ is the penalty parameter.

We can find that by introducing the constraint ${\kern 1pt} {\kern 1pt} {\kern 1pt} {\kern 1pt} \sum\limits_{k = 1,k\neq j}^{{n_i}} {S_{j,k}^i}  = 1,S_{j,k}^i \ge 0$, model (6) treats all classes equally in preserving their own nearest neighbor structures and avoids the trivial solution to weight graph $S$ \cite{li2017locality}. In addition, all weights are adaptively learned from the latent discriminant subspace rather than the complex original space, which enables the method to capture the intrinsic nearest neighbor relationships of samples.

As analyzed in the previous section and proved in many references \cite{zhang2015retargeted,wang2016msdlsr,xiang2012discriminative,zhang2017marginal}, exploiting a more flexible target matrix with large margins between the incorrect and correct classes is very beneficial to improve the discriminability of the projection. Owing to the flexibility and simplicity of the retargeted learning approach \cite{zhang2015retargeted}, we exploit it to replace the strict zero-one target matrix and rewrite our model as follows:
\begin{equation}\label{6}
\small
\begin{split}
\mathop {\min }\limits_{W,S,T} \|T - X^TW\|_F^2 + {\lambda _1}\sum\limits_{i = 1}^C {{n_i}{\sum\limits_{k,j = 1,k\neq j}^{{n_i}} {{S{{_{j,k}^i}^2}}\|W^Tx_j^i - W^Tx_k^i\|_2^2} } } \\
s.t.{\kern 1pt} {\kern 1pt} {\kern 1pt} {\kern 1pt} {T_{i,{l_i}}} - \mathop {\max }\limits_{j \ne {l_i}} {T_{i,j}} \ge 1,\sum\limits_{k = 1,k\neq j}^{{n_i}} {S_{j,k}^i}  = 1,S_{j,k}^i \ge 0
\end{split}
\end{equation}
where $T \in R^{n \times C}$ is the target matrix, $l_i \in \{1,...,C\}$ is the true label index of the $i$th sample.

In most cases, data acquired in the real-world applications usually have high dimensionality and many redundant features even noises. These redundant features or noises residing in the original data are useless even harmful to model training. To address the issue, we further impose a row-sparsity norm constraint on the projection and as a result the final model is:
\begin{equation}\label{8}
\small
\begin{split}
\mathop {\min }\limits_{W,S,T} \left\| {T - X^TW} \right\|_F^2  + {\lambda _2}{\left\| {{W}} \right\|_{2,1}}\\
+ {\lambda _1}\sum\limits_{i = 1}^C {{n_i}{\sum\limits_{k,j = 1, k\neq j}^{{n_i}} {{S{{_{j,k}^i}^2}}\left\| {W^Tx_j^i - W^Tx_k^i} \right\|_2^2} } }\\
s.t.{\kern 1pt} {\kern 1pt} {\kern 1pt} {\kern 1pt} {T_{i,{l_i}}} - \mathop {\max }\limits_{j \ne {l_i}} {T_{i,j}} \ge 1,\sum\limits_{k = 1,k\neq j}^{{n_i}} {S_{jk}^i}  = 1,S_{jk}^i \ge 0
\end{split}
\end{equation}
where $\lambda _2$ is also a penalty parameter.

\emph{\textbf{Proposition 1}}: Introducing the constraint ${\lambda _2}{\left\| W \right\|_{2,1}}$ allows the learned projection $W$ to simultaneously perform feature selection and feature extraction.

\emph{\textbf{Proof}}: As presented in Section II, $\left\| W \right\|_{2,1}$ is defined as ${\left\| W \right\|_{2,1}} = \sum\limits_{i = 1}^m {\sqrt {\sum\limits_{j = 1}^C {W_{i,j}^2} } } $, which is obviously equivalent to the $l_1$ norm constraint ${\left\| a \right\|_1} = \sum\limits_{i = 1}^m {{a_i}} $, where ${a_i} = {\left\| {{W_{i,:}}} \right\|_2} = \sqrt {\sum\limits_{j = 1}^C {W_{i,j}^2} } $. According to the theory of sparse representation, minimizing the optimization problem constrained by the $l_1$ norm will enforce some elements of the corresponding vector to zero \cite{xiang2012discriminative,yan2017protein}. In other words, some elements of vector $a$ will be enforced to zero. Accordingly, if element $a_i$ is enforced to zero, we can deduce that all elements corresponding to the $i$th row of matrix $W$ will be assigned as zero because ${\left\| {{W_{i,:}}} \right\|_2} = \sqrt {W_{i,1}^2 +  \ldots  + W_{i,C}^2}  = 0$. This demonstrates that minimizing problem (8) will adaptively enforce some rows of the projection $W$ to zero. According to the basic rule of matrix multiplication, if all elements of the $i$th row of projection $W$ are zero, then the corresponding $i$th feature of any sample will make no contribution to the generation of the new features during the linear combination. That is to say, these features corresponding to the rows of matrix $W$ with all zero values are not selected during the feature extraction. So we conclude that introducing the sparse constraint ${\lambda _2}{\left\| W \right\|_{2,1}}$ to the model allows the method to earn the feature selection property. Thus we complete the proof.

Proposition 1 allows the method to select the important features for classification and eliminate the negative influence of noises or redundant features. Besides, imposing the $l_{2,1}$ norm on the projection can improve its interpretability \cite{shi2014face}.
\subsection{Solution to the proposed method}
It is obvious that problem (8) has no analytical solution since it contains three variables, \emph{i.e.}, $W,S,T$, in one problem. In this section, we provide an efficient iterative algorithm to obtain their local optimal solutions as follows.

\textbf{Step 1: Calculate variable $W$.} For convenience ,we define
\begin{equation}\label{9}
\begin{split}
L(W) = \left\| {T - X^TW} \right\|_F^2 + {\lambda _2}{\left\| {{W}} \right\|_{2,1}}\\
+ {\lambda _1}\sum\limits_{i = 1}^C {{n_i}{\sum\limits_{k,j = 1,k\neq j}^{{n_i}} {{S{{_{j,k}^i}^2}}\left\| {W^Tx_j^i - W^Tx_k^i} \right\|_2^2} } }
\end{split}
\end{equation}

Problem (9) can be simplified as follows:
\begin{equation}\label{10}
\small
L(W) = \left\| {T - X^TW} \right\|_F^2 + {\lambda _1}Tr\left( {W^T{S_W}W} \right) + {\lambda _2}{\left\| {{W}} \right\|_{2,1}}
\end{equation}
where ${S_W} = \sum\limits_{i = 1}^C {{n_i} {\sum\limits_{k,j = 1,k\neq j}^{{n_i}} {{S{{_{j,k}^i}^2}}\left( {x_j^i - x_k^i} \right){{\left( {x_j^i - x_k^i} \right)}^T}} } }$. Then we can obtain the optimal $W$ by setting the derivative of $L(W)$ with respect to $W$ to zero as follows:
\begin{equation}\label{11}
\begin{split}
X\left( {{X^T}W - T} \right) + {\lambda _1}{S_W}W + \frac{{{\lambda _2}}}{2}DW = 0\\
\Rightarrow W = {\left( {X{X^T} + {\lambda _1}{S_W} + \frac{{{\lambda _2}}}{2}D} \right)^{ - 1}}XT
\end{split}
\end{equation}
where $D \in R^{m \times m}$ is a diagonal matrix and its each diagonal element ${D_{i,i}} = {1 \mathord{\left/
 {\vphantom {1 {\sqrt {\sum\limits_{j = 1}^C {w_{i,j}^2} } }}} \right.
 \kern-\nulldelimiterspace} {\sqrt {\sum\limits_{j = 1}^C {W_{i,j}^2} } }}$ \cite{xiang2012discriminative}.

\textbf{Step 2: Calculate variable $S$.} Fixing variables $T$ and $W$, variable $S$ can be obtained by solving the following problem:
\begin{equation}\label{13}
\begin{split}
\mathop {\min }\limits_S \sum\limits_{i = 1}^C {{n_i} {\sum\limits_{k,j = 1,k\neq j}^{{n_i}} {S{{_{j,k}^i}^2}\left\| {{W^T}x_j^i - {W^T}x_k^i} \right\|_2^2} } } {\kern 2pt}  \\
s.t. {\kern 2pt}  \sum\limits_{k = 1,k\neq j}^{{n_i}} {S_{j,k}^i}  = 1,S_{j,k}^i \ge 0
\end{split}
\end{equation}

Problem (12) can be further simplified into the following subproblems:
\begin{equation}\label{14}
\mathop {\min }\limits_{{\sum\limits_{k = 1,k\neq j}^{{n_i}} {S_{j,k}^i}  = 1,S_{j,k}^i \ge 0}} \sum\limits_{k = 1,k\neq j}^{{n_i}} {S{{_{j,k}^i}^2}\left\| {{W^T}x_j^i - {W^T}x_k^i} \right\|_2^2}
\end{equation}

\textbf{\emph{Theorem 1}} \cite{li2017locality}. Given a positive vector $b\in R^{1\times n}$ ($\forall i$ ($1 \le i \le n$), ${b_i} > 0$), problem $\mathop {\min }\limits_{\sum\limits_{i = 1}^n {{a_i}}  = 1,{a_i} \ge 0} \sum\limits_{i = 1}^n {a_i^2{b_i}} $ has the optimal solution as ${a_i} = \frac{1}{{{b_i}}}{\left( {\sum\limits_{k = 1}^n {\frac{1}{{{b_k}}}} } \right)^{ - 1}}$.

\emph{\textbf{Proof}}. The detailed proof process is moved to the Appendix A in the supplementary material.

According to \textbf{Theorem 1}, we can obtain the optimal solution to problem (13):
\begin{equation}\label{15}
\small
S_{j,k}^i = \frac{1}{{\left\| {{W^T}x_j^i - {W^T}x_k^i} \right\|_2^2}}{\left( {\sum\limits_{p = 1,p\neq j}^{{n_i}} {\frac{1}{{\left\| {{W^T}x_j^i - {W^T}x_p^i} \right\|_2^2}}} } \right)^{ - 1}}
\end{equation}

\textbf{Step 3: Calculate variable $T$.} Fixing variables $W$ and $S$, the subproblem to variable $T$ is as follows:
\begin{equation}\label{16}
\mathop {\min }\limits_{{T_{i,{l_i}}} - \mathop {\max }\limits_{j \ne {l_i}} {T_{i,j}} \ge 1} \left\| {T - {X^T}W} \right\|_F^2
\end{equation}

Problem (15) is a typical constrained quadratic programming problem \cite{koyejo2013retargeted}.

\textbf{\emph{Theorem 2}} \cite{zhang2015retargeted}. For any given vector $g = \left[ {{g_1},{g_2}, \ldots ,{g_n}} \right]$, the optimal solution of problem $\mathop {\min }\limits_{{t_h} - \mathop {\max }\limits_{i \ne h} {t_i}} \left\| {t - g} \right\|_2^2$ is
\begin{equation}\label{17}
{t_i} = \left\{ {\begin{array}{*{20}{c}}
{{g_i} + \Delta ,}&{if{\kern 1pt} {\kern 1pt} i = h}\\
{{g_i} + \min \left( {\Delta  - {v_i},0} \right),}&{otherwise}
\end{array}} \right.
\end{equation}
where ${v_i} = 1 + {g_i} - {g_h}$, $\Delta  = \frac{{\sum\limits_{i \ne h} {{v_i}\Upsilon \left( {\Gamma '\left( {{v_i}} \right),0} \right)} }}{{1 + \sum\limits_{i \ne h} {\Upsilon \left( {\Gamma '\left( {{v_i}} \right),0} \right)} }}$. $\Upsilon \left(  \cdot  \right)$ is a logic function and has the following defination $\Upsilon \left( {a,b} \right) = \left\{ {\begin{array}{*{20}{c}}
{1,}&{if{\kern 1pt} {\kern 1pt} a > b}\\
{0,}&{otherwise}
\end{array}} \right.$. Function $\Gamma '\left( x \right) = 2(x + \sum\limits_{i \ne h} {\min \left( {x - {v_i},0} \right)} )$.

\emph{\textbf{Proof}}: Please refer to Appendix B in supplementary material for the completed proof process.

It is obvious that problem (15) can be decomposed into the following subproblems
\begin{equation}\label{17}
\mathop {\min }\limits_{{T_{i,{l_i}}} - \mathop {\max }\limits_{j \ne {l_i}} {T_{i,j}} \ge 1} \left\| {{T_{i,:}} - {G_{i,:}}} \right\|_2^2
\end{equation}

According to \textbf{Theorem 2}, we can fast obtain the optimal solution $T_{i,:}$ of problem (17). By computing all rows of $T$ separately according to (17) and \textbf{Theorem 2}, the optimal solution $T$ can be finally obtained.

Algorithm 1 summarizes the completed optimization steps of our method.
\begin{algorithm}[htb]
 \caption{: ALPR (solving problem (8))}
  \begin{algorithmic}
  \STATE {\textbf{Input:} Data matrix $X \in {R^{m \times n}}$, label matrix $Y \in {R^{n \times C}}$, class number $C$, parameters ${\lambda _1},{\lambda _2}$.
   \\\textbf{Initialization:} Matrix $W \in {R^{m \times C}}$ with random values; $T=Y$, Graph $S$ is constructed as the \emph{k}-nearest neighbor graph with binary values.}
  \WHILE{not converged}{\STATE{1. Update $W$ by using (11);}\\
{2. Update all elements of $S$ by using (14);}\\
{3. Update all rows of $T$ according to Theorem 2.}\\
}
\ENDWHILE
\STATE {\textbf{Output:} $W,S,T$}
\end{algorithmic}
\end{algorithm}
\subsection{Classification based on ALPR}
After obtaining the projection $W$, LR based methods transform all the original data into the discriminant subspace via $W^{T}X$ for the subsequent classification. It is undoubted that each projected sample can be viewed as its new representation, which is obtained by the weighted linear combination of projection $W$ and all original features of the corresponding sample. However, this simple weighted combination approach may magnify the negative influence of noises or redundant features residing in the high-dimensional data. Thanks to the feature selection property of the used $l_{2,1}$ norm constraint, our method has the potential to discover the discriminability of different features. As analyzed in \emph{Proposition 1}, by imposing $l_{2,1}$ norm on the projection, some rows of projection $W$ corresponding to the unimportant features will be adaptively assigned very small values even 0. Meanwhile, these rows also have very small $l_2$ norm values. This illustrates that the discriminability of each feature can be commonly measured by the $l_2$ norm of the corresponding row of projection $W$ \cite{xiang2012discriminative}. Thus we can enforce those rows to zero directly to further eliminate the negative influence of noises and redundant features. Inspired by this motivation, we provide the following approach listed in Algorithm 2 to implement the classification, in which a threshold value $\rho$ is set to select those discriminative features.
 \begin{algorithm}[htb]
 \caption{: Classification based on ALPR}
  \begin{algorithmic}
  \STATE {\textbf{Input:} Training data $X \in {R^{m \times n}}$ and label $L \in {R^n}$, projection $W \in R^{m \times c}$, test sample $a \in R^m$, threshold value $\rho$.
 \begin{itemize}
   \item Obtain the discriminability $p_i$ of each feature by calculating the $l_2$ norm of the corresponding row vector of $W$;
   \item Force all elements of the rows of the projection $W$ that satisfy $p_i < \rho$ to zero.
   \item Project all training sample and test sample into the target space via the new projection $W$ and use the nearest neighbor classify to recognize the test sample.
 \end{itemize}}
\STATE {\textbf{Output:} The label of the test sample.}
\end{algorithmic}
\end{algorithm}

\section{Analysis of the proposed method}
\subsection{Computational complexity}
In this section, we mainly analyze the computational complexity of the proposed optimization algorithm listed in Algorithm 1. For simplicity, we do not consider the computational complexities of some simple matrix operations, such as matrix addition, subtraction, multiplication, and element-wise based matrix division since these operations can be efficiently computed. Overall, there are three main variables, \emph{i.e.}, $W$, $S$, and $T$, to calculate from Algorithm 1. In Step 1, \emph{i.e.}, updating variable $W$, the major computational cost is the matrix inverse operation which has the computational complexity of $O\left( {{m^3}} \right)$ for an $m \times m$ matrix \cite{lu2018structurally}. Thus the total computational complexity of Step 1 is about $O\left( {{m^3}} \right)$. For Steps 2 and 3, it is obvious that these two steps only contain some simple matrix operations and thus their computational complexities can be ignored. In summary, the total computational complexity of the optimization approach listed in Algorithm 1 is about $O\left( {\tau {m^3}} \right)$, where $\tau $ is the iteration number.

\subsection{Convergence analysis}
From the previous presentation, all variables can be simply calculated with the closed form solutions in their own subproblems. Meanwhile, we can prove the following proposition.

\textbf{\emph{Proposition 2}}: For the optimization problem (8), each subproblem is convex with respect to variables $W,S,T$, respectively.

\textbf{\emph{Proof:}} From (9), the subproblem to variable $W$ is a typical sparse constraint optimization problem. Because the sparse ${l_{2,1}}$ norm is a convex function [17], thus problem (9) is a convex optimization subproblem. For variable $S$, it is obvious that all constraints with respect to variable $S$ are convex, and thus subproblem (12) is also convex [32, 33]. The subproblem (16) with respect to variable $T$ is also a typical convex problem, \emph{i.e.}, convex constraint quadratic programming problem [34]. Therefore, all subproblems with respect to variables $W,S,T$ are convex, respectively. Thus we complete the proof.

Based on the proposition 2, we can derive the following Theorem 3 to the presented Algorithm 1 [19, 32].

\textbf{\emph{Theorem 3}}: The optimization approach presented in Algorithm 1 monotonically decreases the objective value of problem (8).

\textbf{\emph{Proof}}: Let $L\left( {{W^t},{S^t},{T^t}} \right)$ be the objective function value of problem (8) at the $t$th iteration. At the $\left( {t + 1} \right)$th iteration, the optimal solution of ${W^{t + 1}}$ is first calculated by solving the subproblem (9). Because subproblem (9) is convex, thus we have the following inequation after this iteration step
\begin{equation}\label{15}
L\left( {{W^{t + 1}},{S^t},{T^t}} \right) \le L\left( {{W^t},{S^t},{T^t}} \right)
\end{equation}

Similarly, owing to the convex property of subproblems to variables $W$ and $T$, we can obtain the following inequations after the corresponding iterations:
\begin{equation}\label{15}
L\left( {{W^{t + 1}},{S^{t + 1}},{T^t}} \right) \le L\left( {{W^{t + 1}},{S^t},{T^t}} \right)
\end{equation}
\begin{equation}\label{15}
L\left( {{W^{t + 1}},{S^{t + 1}},{T^{t + 1}}} \right) \le L\left( {{W^{t + 1}},{S^{t + 1}},{T^t}} \right)
\end{equation}

Combing Eqs. (22), (23) and (24), we can obtain
\begin{equation}\label{15}
L\left( {{W^{t + 1}},{S^{t + 1}},{T^{t + 1}}} \right) \le L\left( {{W^t},{S^t},{T^t}} \right)
\end{equation}

Therefore, we complete the proof.

\textbf{Theorem 3} provides some assurances to the convergence property of the proposed method. Since the objective function (8) is lower bounded, the proposed method will finally find a local optimal solution that makes the objective function value converge \cite{zhang2018binary}. In the subsequent section, we will further prove its convergence property based on some experiments.
\subsection{Connections to other methods}
In this section, we analyze the connections and differences between the proposed method and some related LR methods, such as ReLSR \cite{zhang2015retargeted}, MSRL \cite{zhang2017marginal}, DLSR \cite{xiang2012discriminative} and SLRR \cite{cai2013equivalent}, etc.

(1) Connections to ReLSR and MSRL: The discriminative regression model of MSRL is as follows:
\begin{equation}\label{15}
\small
\begin{split}
\mathop {\min }\limits_{W,A,B,T,P} \left\| {T - {W^T}X} \right\|_F^2 + \alpha \left\| W \right\|_F^2{\rm{ + }}\beta \left\| {W - AB} \right\|_F^2 \\
+ \lambda \sum\limits_{i,j}^n {\left( {\left\| {{W^T}{x_i} - {W^T}{x_j}} \right\|_2^2{P_{i,j}} + \sigma P_{i,j}^2} \right)} \\
s.t.{\kern 1pt} {\kern 1pt} {T_{i,{l_i}}} - \mathop {\max }\limits_{j \ne {l_i}} {T_{i,j}} \ge 1,{A^T}A = I,0 \le {P_{i,j}} \le 1,P\textbf{\emph{1}} = \textbf{\emph{1}}
\end{split}
\end{equation}

From the models of MSRL in (22) and ReLSR in (2), we can find that the proposed method and MSRL are the two extensions of ReLSR indeed. By introducing some valuable constraints to the model, MSRL and the proposed method can learn a more compact and discriminative projection than ReLSR. Although there are some similar points between MSRL and the proposed method, they still have many differences as follows. (i) MSRL exploits an unsupervised graph regularization term to guide the projection learning. While in our method, the graph regularization term is supervised. For MSRL, the unsupervised graph regularization term is not perfect and has many shortcomings. For example, in MSRL, samples from different classes may be regarded as the nearest neighbor samples and connected with large weight, which is obviously unreasonable. Besides, MSRL is sensitive to the number of nearest neighbors. Compared with MSRL, our method does not have the above problems since it can construct a more reliable graph by introducing the novel supervised graph learning term. (ii) MSRL imposes `Frobenius norm' on the projection matrix. In our method, the sparse ${l_{2,1}}$ norm is introduced. Compared with MSRL, the proposed method has the feature selection property, which is able to select the most important features for regression and has the potential to reduce the negative influence of the redundant features and noises. (iii) We can find that the proposed method has less penalty parameters than MSRL, which greatly reduces the complexity of parameter selection. In Fig.2, we have plotted the first 100 rows of the projections learned by ReLSR, MSRL, and the proposed method on the PIE database. From Fig.2, it is obvious that some rows of the projection (Fig.2(c)) learned by the proposed method are adaptively enforced to zero while the other two projections do not show this phenomenon. As analyzed in proposition 1, some features of original data corresponding to these rows will not be selected for feature extraction. In Fig.3(a), we have marked these unselected features with `black point' on the original images. From Fig.3(a), we can find that some similar areas among different faces are marked while features of the remaining areas such as eyes, mouth, and nose, etc. are treated as important features. This also proves that the proposed method has the potential to select those important features from the original high-dimensional data. Meanwhile, in Fig.3(b), we have plotted the weight graph of features, in which the weight corresponding to the $i$th feature is calculated as $||W_{i,:}||_2$ \cite{xiang2012discriminative}. From Fig.3(b), it is easy to discover the importance degrees of different features in classification. This demonstrates that projection learned by our method has good interpretability for features. In summary, the proposed method has many superior properties in comparison with MSRL and ReLSR, which allow it to perform better than the other methods.

(2) Connections to SLRR and DLSR. Similar to the proposed method, SLRR and DLSR all impose the sparse ${l_{2,1}}$ norm constraint on the projection matrix. The objective functions of SLRR and DLSR are respectively formulated as follows
\begin{equation}\label{15}
\mathop {\min }\limits_{rank\left( W \right) \le s} \left\| {Y - {X^T}W} \right\|_F^2 + \lambda {\left\| W \right\|_{2,1}}
\end{equation}
\begin{equation}\label{15}
\mathop {\min }\limits_{W,M \ge 0} {\left\| {Y - B \odot M - {X^T}W} \right\|_{2,1}} + \lambda {\left\| W \right\|_{2,1}}
\end{equation}
where $B$ is a constant matrix and defined as follows
\begin{equation}\label{15}
{B_{i,j}} = \left\{ {\begin{array}{*{20}{c}}
{1,}&{if{\kern 1pt} {\kern 1pt} {l_i} = j}\\
{ - 1,}&{otherwise}
\end{array}} \right.
\end{equation}
where $l_i$ is the true class index of sample $x_i$.

From (23), (24), and (8), the proposed method and DLSR all exploit the target learning technique to enlarge the margins of samples from different classes, thus can learn a more discriminative projection than SLRR. Compared with DLSR, our method not only exploits a more flexible target learning technique, but also introduces an adaptive graph regularization term to further improve the compactness of the projection. These properties encourage the proposed method to learn a more discriminative projection than DLSR and avoid overfitting, and thus can obtain a better performance.
\begin{figure}[htb]
\centering
\subfloat[ReLSR]{\includegraphics[width=1in]{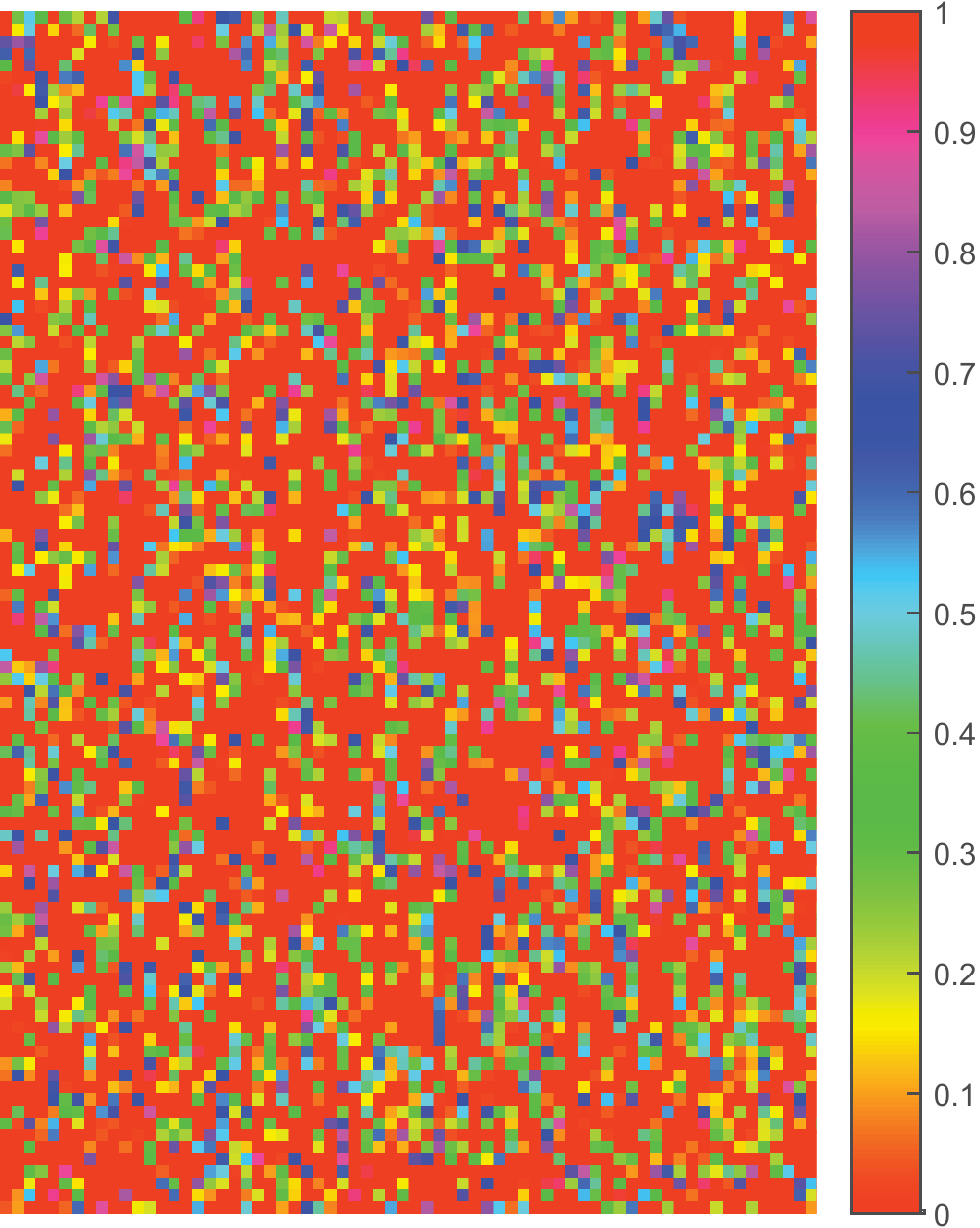}
\label{fig_first_case}}
\hfil
\subfloat[MSRL]{\includegraphics[width=1in]{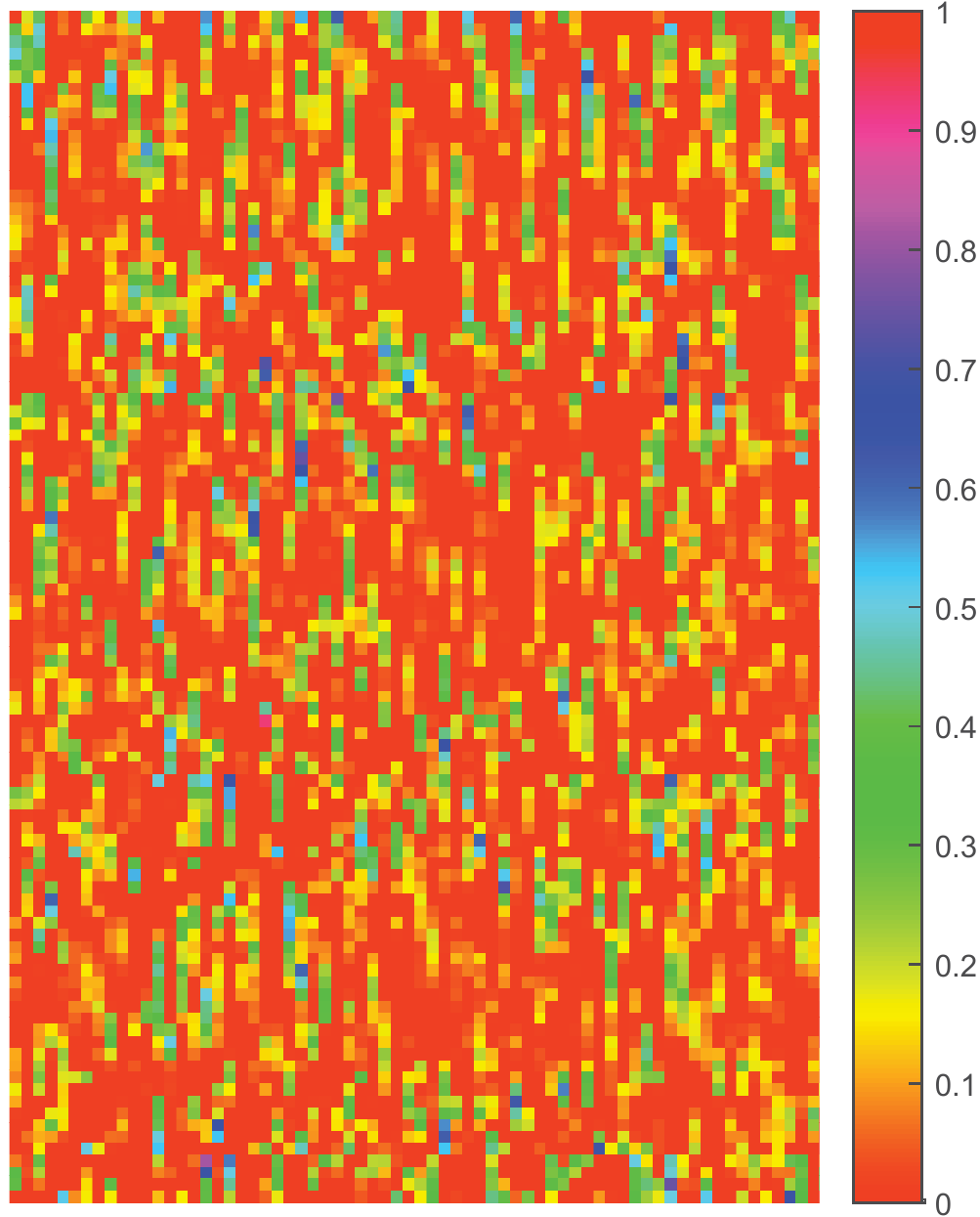}
\label{fig_second_case}}
\hfil
\subfloat[ALPR]{\includegraphics[width=1in]{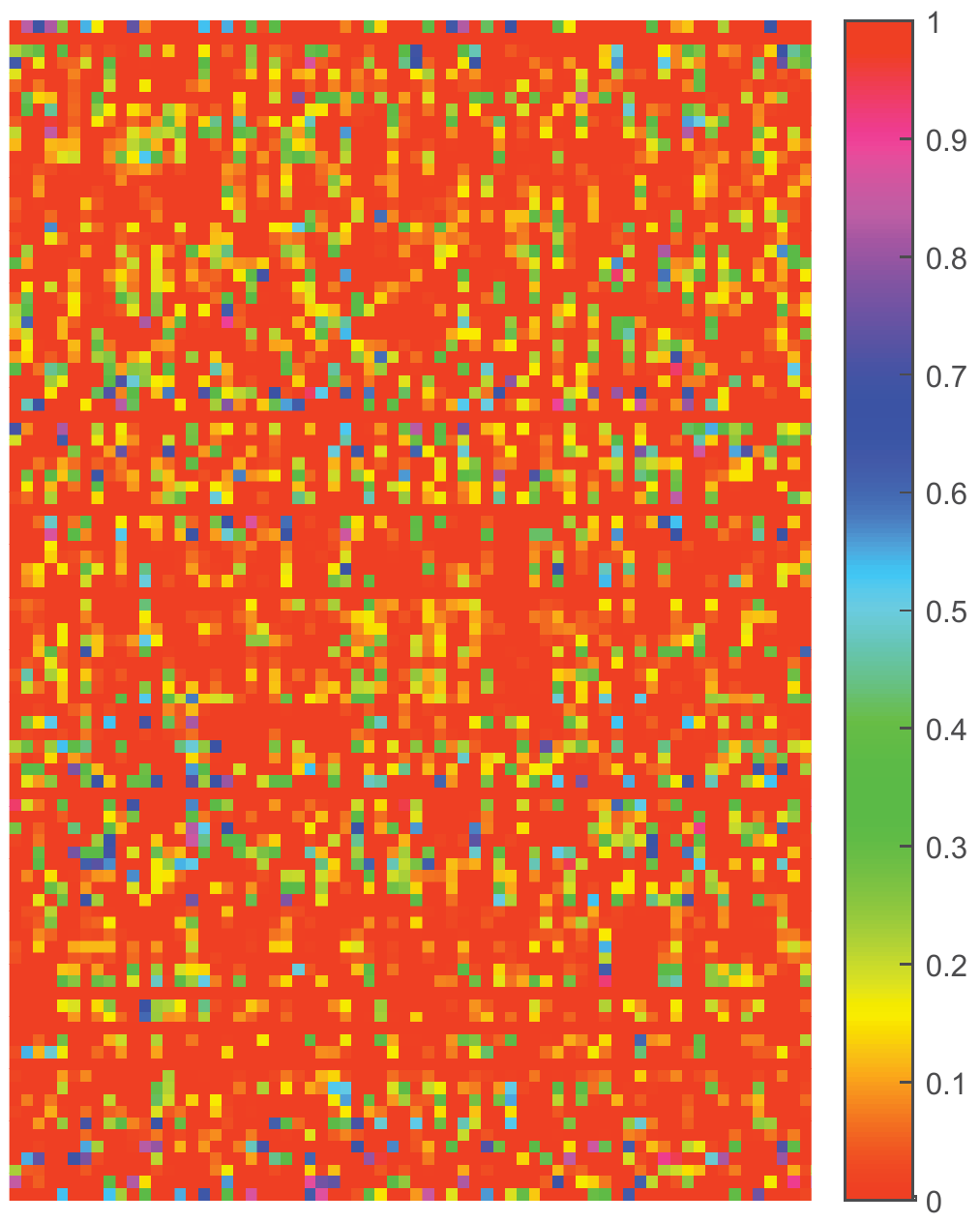}
\label{fig_second_case}}
\caption{Projections learned by ReLSR, MSRL, and our method on the PIE database, in which 20 samples per class are randomly selected as training samples. Note: all figures are shown in the HSV color space; for better comparison, we only plot the first 100 rows of these projections. From the colorbar, we can infer all element values of these projections. Moreover, we can clearly see that element values in some rows are 0 or close to 0 in (c).}
\label{fig:3}
\end{figure}
\begin{figure}[htb]
\centering
\subfloat[Original face images]{\includegraphics[width=1.5in]{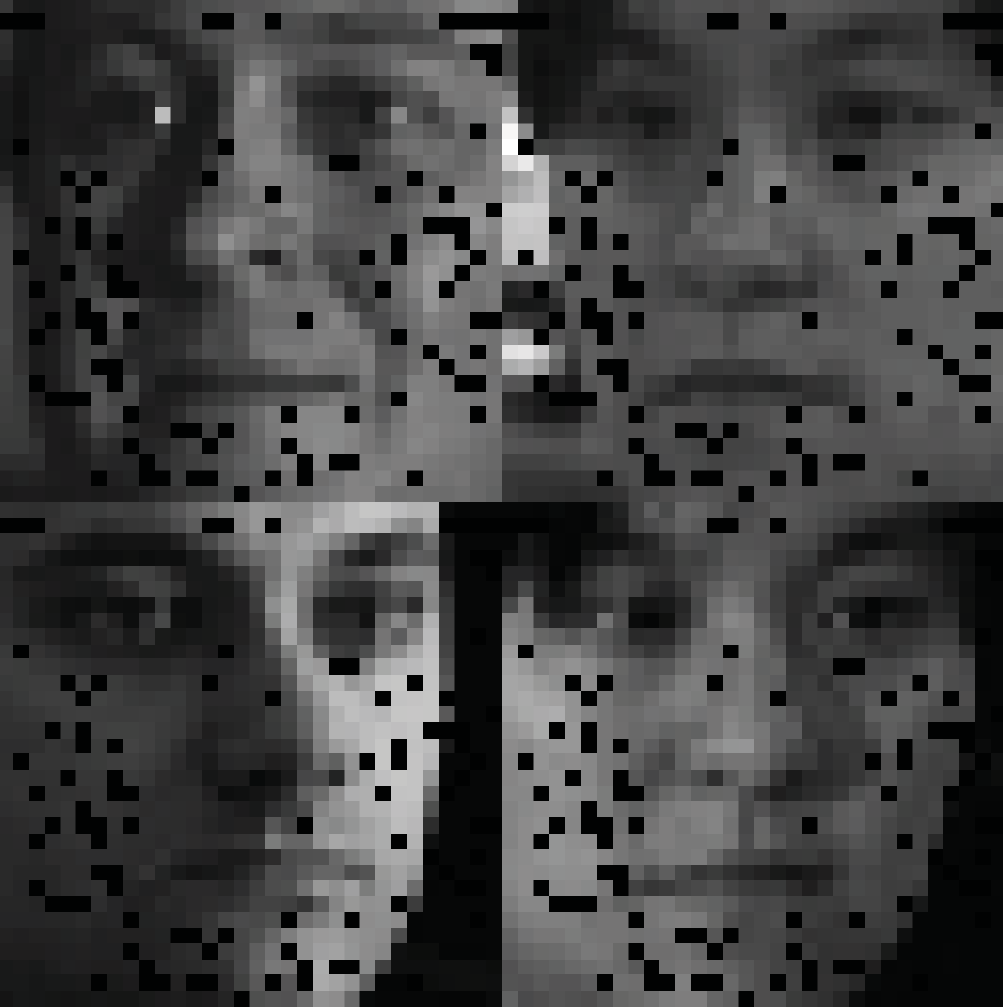}
\label{fig_second_case}}
\hfil
\subfloat[Weight graph of features]{\includegraphics[width=1.5in]{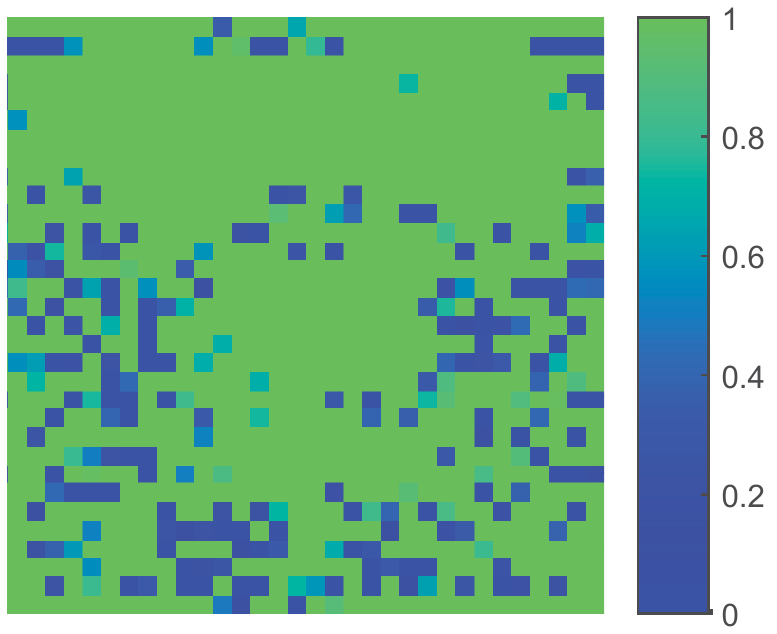}
\label{fig_second_case}}
\caption{The original training samples of the PIE database and the weighted graph of features, in which the `black pixel points' in (a) denote the rows of projection with all zeros of the proposed method. Note: weight values in (b) are normalized to the range of [0,1].}
\label{fig:3}
\end{figure}

\section{Experiments and analysis}
In this section, we evaluate the proposed method on the synthetic database and five real-world databases. Several related LR based methods, including linear regression classification (LRC) \cite{naseem2010linear}, sparse representation based classification (SRC) \cite{wright2009robust}, collaborative representation based classification (CRC) \cite{zhang2012collaborative}, support vector machine (SVM){\footnote{We exploit LibSVM toolbox to implement experiments. LibSVM is available at https://www.csie.ntu.edu.tw/~cjlin/libsvm/.}} \cite{chang2011libsvm}, LRLR \cite{cai2013equivalent}, low-rank ridge regression (LRRR) \cite{cai2013equivalent}, SLRR \cite{cai2013equivalent}, discriminative least squares regression (DLSR) \cite{xiang2012discriminative}, ReLSR \cite{zhang2015retargeted}, DRLS \cite{xue2009discriminatively}, MSRL \cite{zhang2017marginal}, constrained least square regression (CLSR) \cite{yuan2018constrained}, and groupwise retargeted least-squares regression (GReLSR) \cite{wang2018groupwise}, are chosen to compare with the proposed method. Among these methods, LRRR is an extension of LRLR, which imposes the `Frobenius' norm on the low-rank projection. CLSR and GReLSR can be viewed as the extensions of ReLSR, which mainly introduce the group based label relaxation technique to improve the performance. For the proposed method, the threshold value $\rho$ is set to 0.0001 to all databases.

\subsection{Experiments and analysis on the synthetic database}
Preserving the geometric structure is very important to discriminant analysis, especially for some databases with manifold structure. Following the experimental settings in \cite{li2017locality}, we synthesize the typical manifold data, \emph{i.e.}, three-ring data with two different amplitudes of the third feature, to prove the effectiveness of the proposed method in dealing with such type of data. For convenience, we refer to the three-ring data with amplitude $\left[ { - 20,20} \right]$ as Th1, and refer to the other one with amplitude $\left[ { - 2000,2000} \right]$ as Th2. Fig.4 (a) shows the typical data of the Th1, which contains three features. For Th1 and Th2, the first two features are centrally distributed in the circle style, while the third feature can be viewed as the noise to some extent. In our experiments, each synthesized three-ring data is composed of 3 classes and 1000 samples per class, in which 500 samples are randomly selected from each class to form the training set, and the remaining samples are treated as the test set accordingly.

Table I enumerates the experimental results of different methods including the nearest neighbor classifier (NC) on Th1 and Th2. In Fig.4, we plot the projected test samples and their predicted labels of different methods on Th1, and show the projection learned by our method on Th1. From the experimental results in Table I, it is obvious that many methods except SVM and the proposed method perform worse than the simplest classifier, \emph{i.e.}, NC, on Th1 and Th2. In addition, from Fig.4, we can also find that only the proposed method can simultaneously preserve the intrinsic structure well and obtain the satisfactory classification result. Although LRLR, LRRR, and SLRR can preserve the similar structure as the original data, they cannot predict these test samples correctly. Therefore, these experimental results prove the effectiveness of the proposed method in classifying the data with manifold structure. Moreover, from Table I, we can also find that with the amplitude of the third feature (noise) increasing, the classification accuracies of almost all methods decrease dramatically. And it is obvious that the classification accuracy of our method only decreases about 0.06\% when the noise increases. These demonstrate that the proposed method is superior to the other methods for the classification tasks with noises. Furthermore, from Fig.4 (l), we can find that the feature extraction weights corresponding to the third feature (noise) of Th1 are adaptively set as 0. This proves that introducing the row-sparsity norm constraint is valuable, which can effectively reduce the negative influence of noises. In summary, the proposed method is not only suitable to classify the data with manifold structure, but also robust to noise to some extent.
\begin{figure*}
\centering
\subfloat[Test data of Th1]{\includegraphics[width=1.1in]{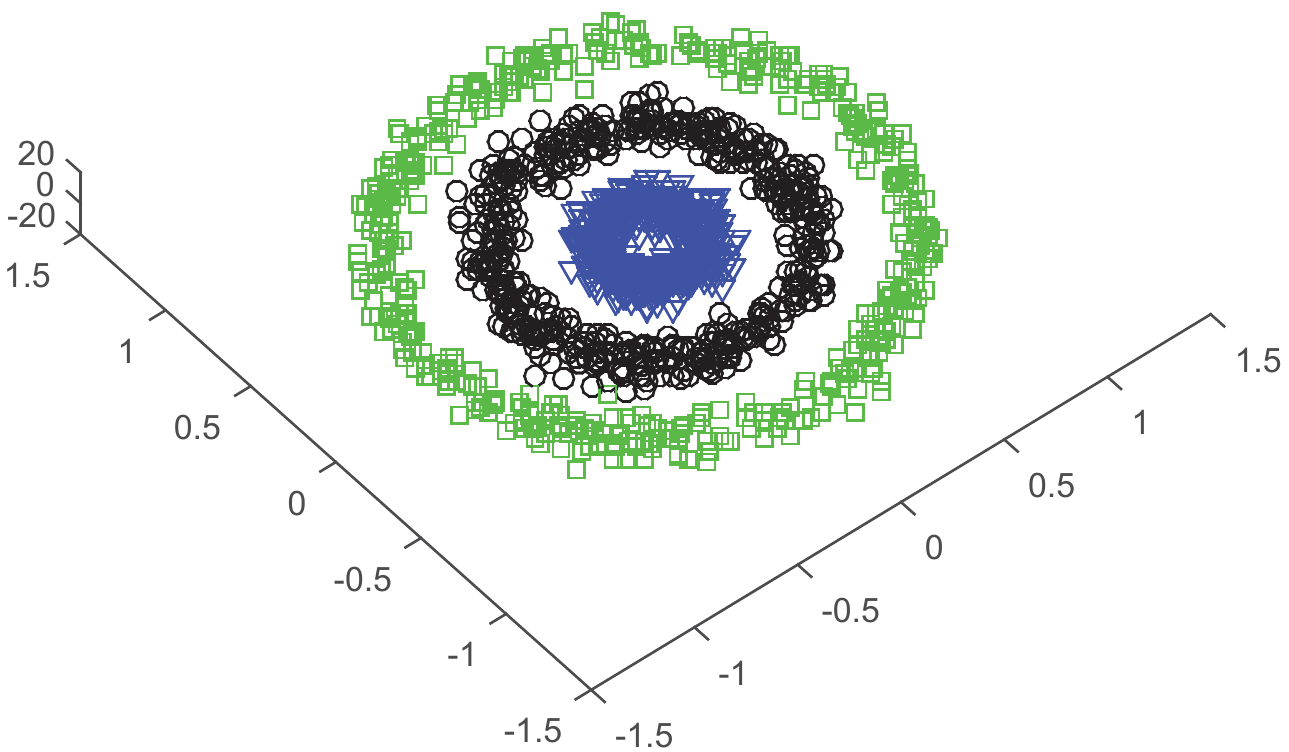}
\label{fig_first_case}}
\hfil
\subfloat[LRLR]{\includegraphics[width=1.1in]{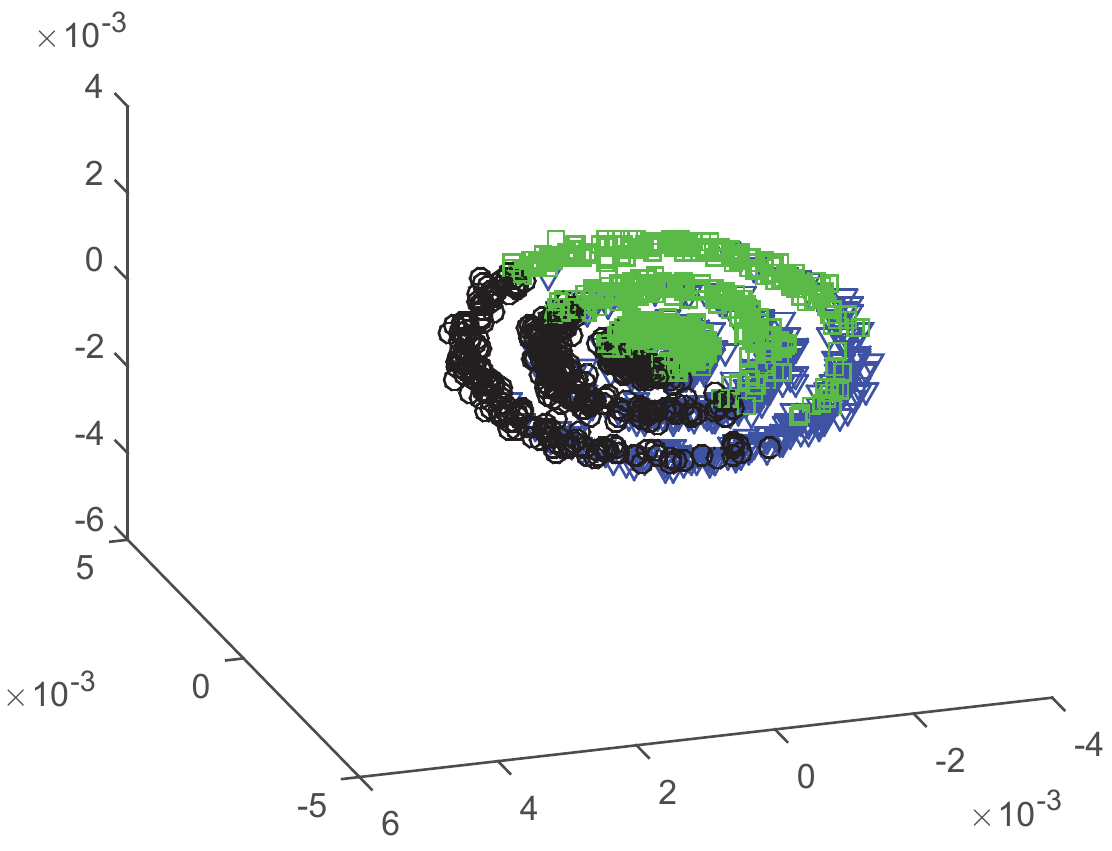}
\label{fig_second_case}}
\hfil
\subfloat[LRRR]{\includegraphics[width=1.1in]{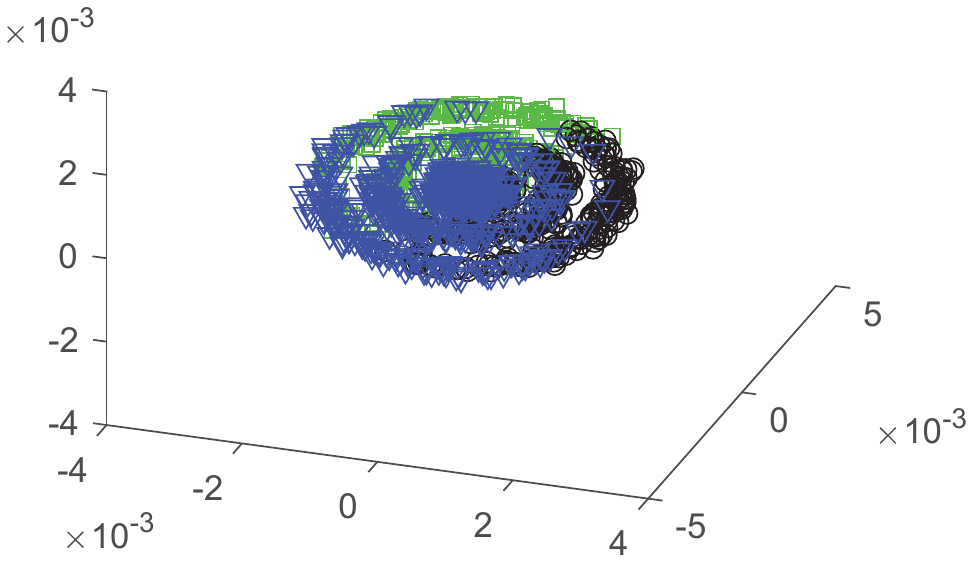}
\label{fig_second_case}}
\hfil
\subfloat[SLRR]{\includegraphics[width=1.1in]{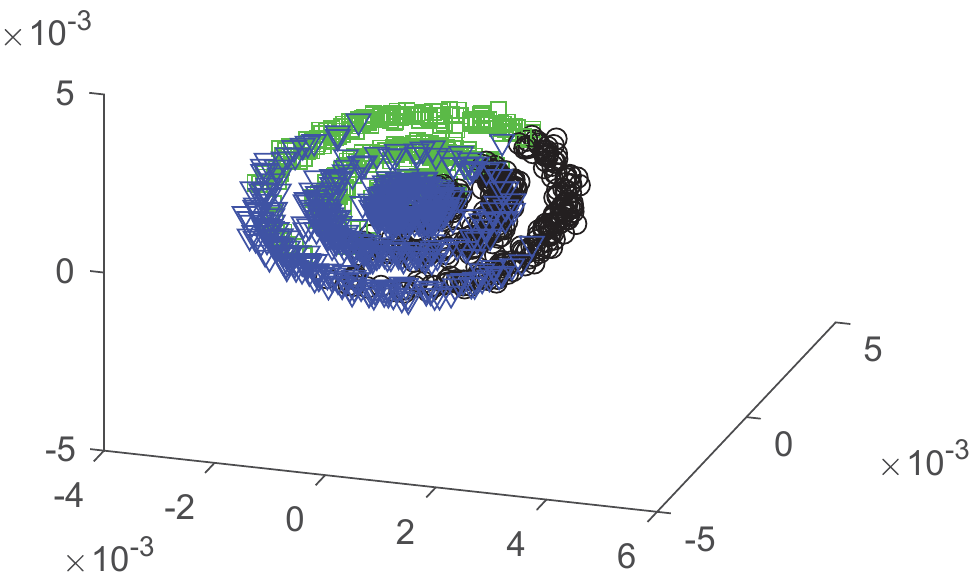}
\label{fig_second_case}}
\hfil
\subfloat[DLSR]{\includegraphics[width=1.1in]{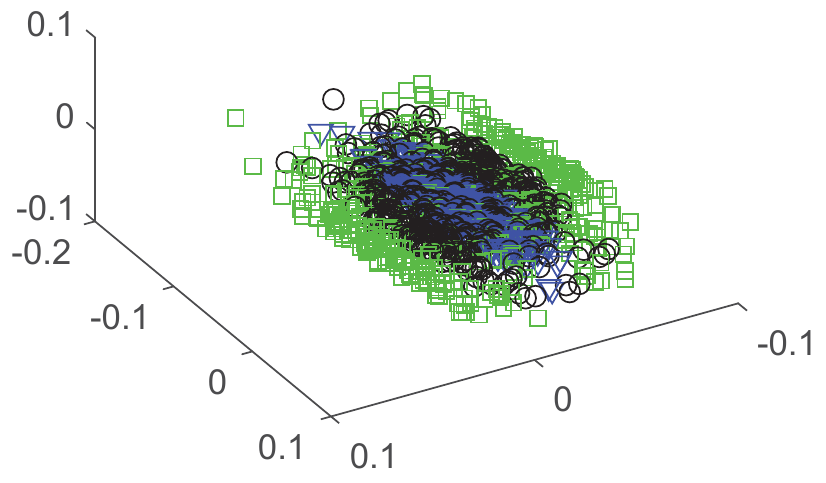}
\label{fig_second_case}}
\hfil
\subfloat[ReLSR]{\includegraphics[width=1.1in]{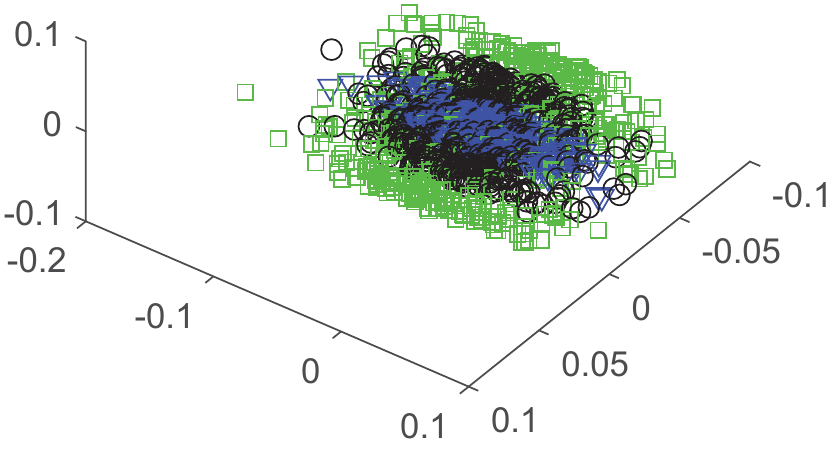}
\label{fig_second_case}}
\hfil
\subfloat[DRLS]{\includegraphics[width=1.1in]{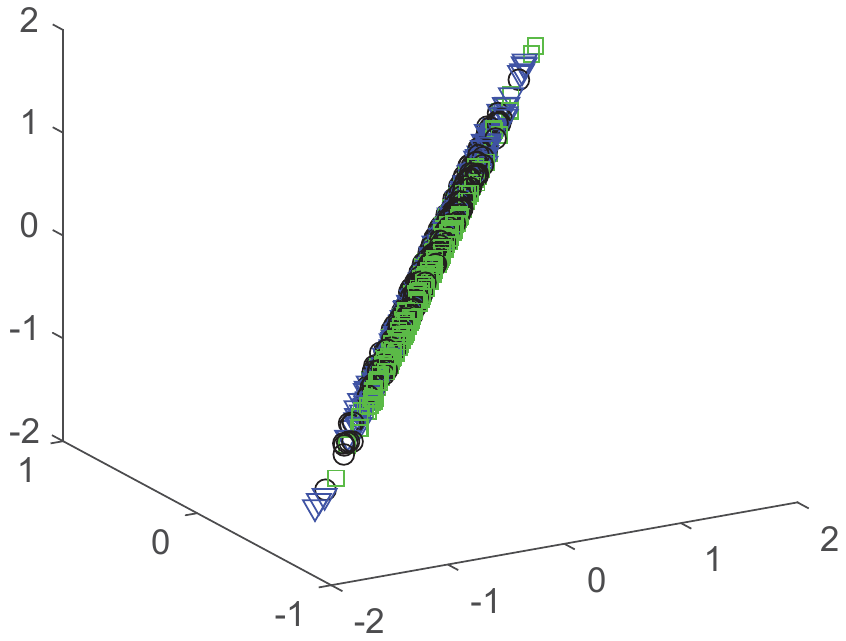}
\label{fig_second_case}}
\hfil
\subfloat[MSRL]{\includegraphics[width=1.1in]{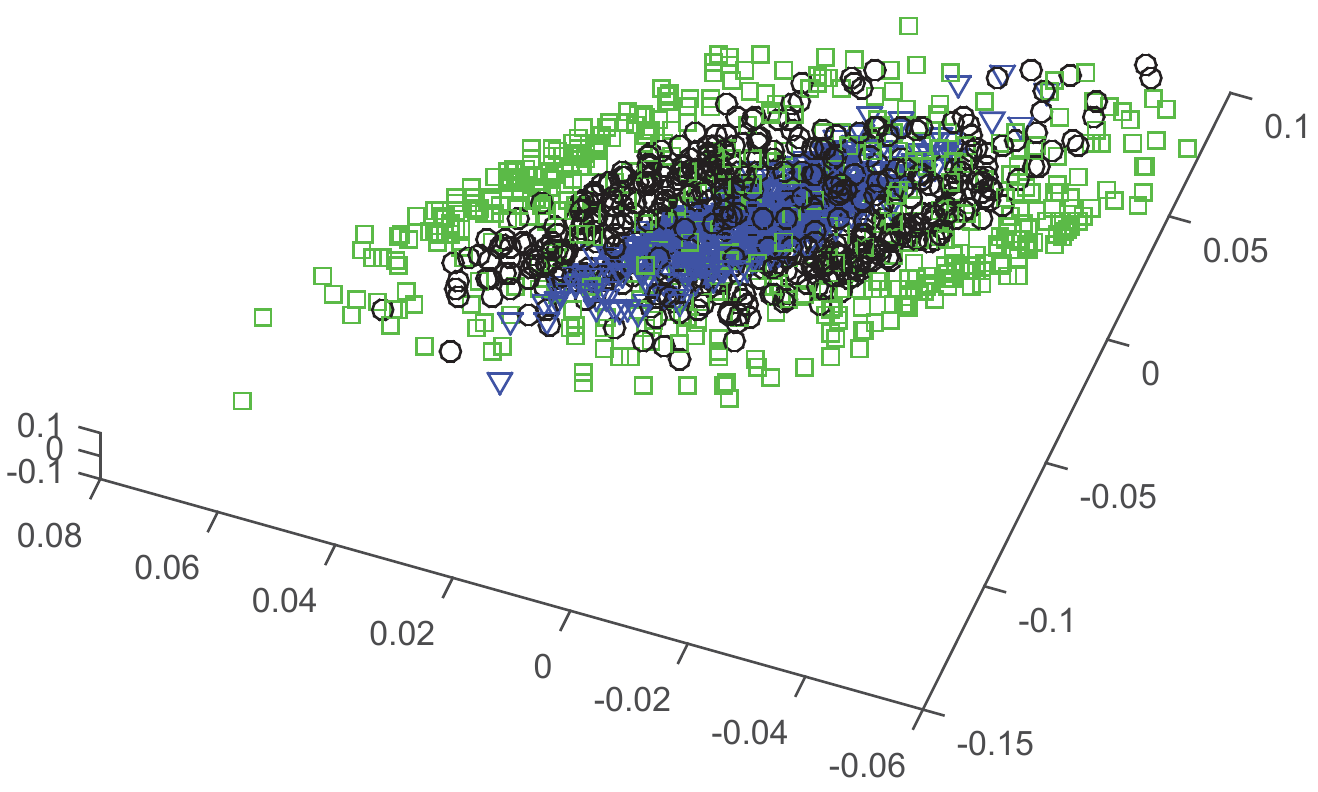}
\label{fig_second_case}}
\hfil
\subfloat[CLSR]{\includegraphics[width=1.1in]{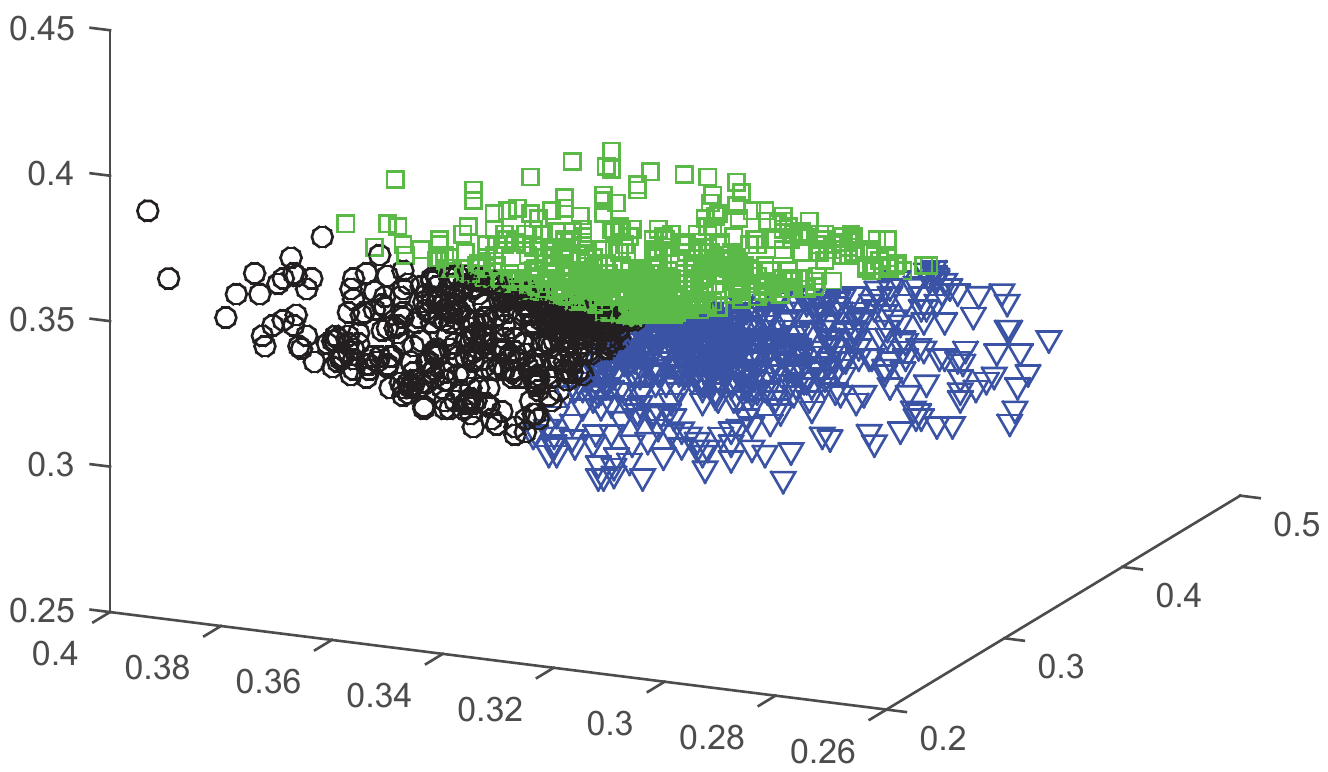}
\label{fig_second_case}}
\hfil
\subfloat[GReLSR]{\includegraphics[width=1.1in]{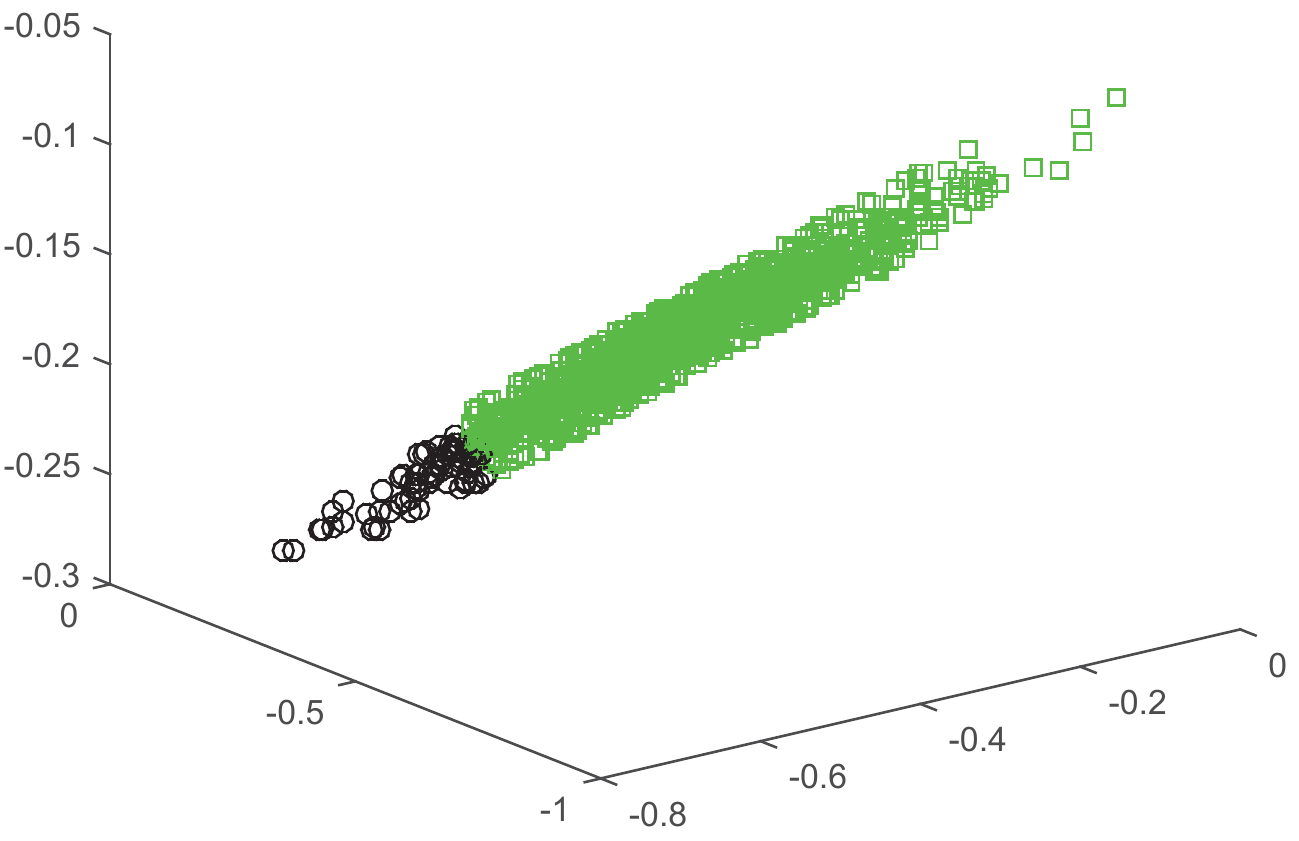}
\label{fig_second_case}}
\hfil
\subfloat[ALPR]{\includegraphics[width=1.1in]{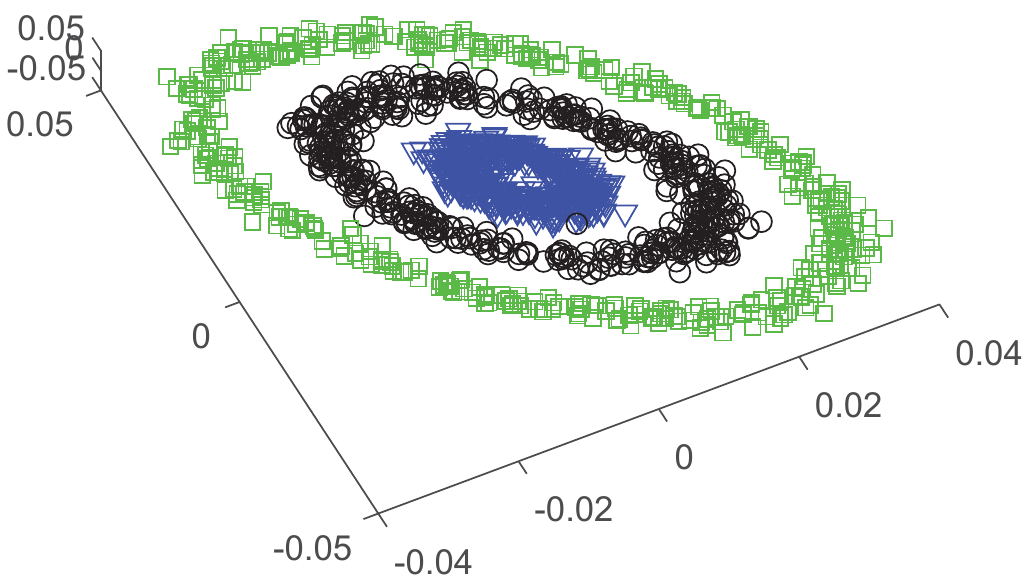}
\label{fig_second_case}}
\hfil
\subfloat[Projection of ALPR]{\includegraphics[width=1.1in]{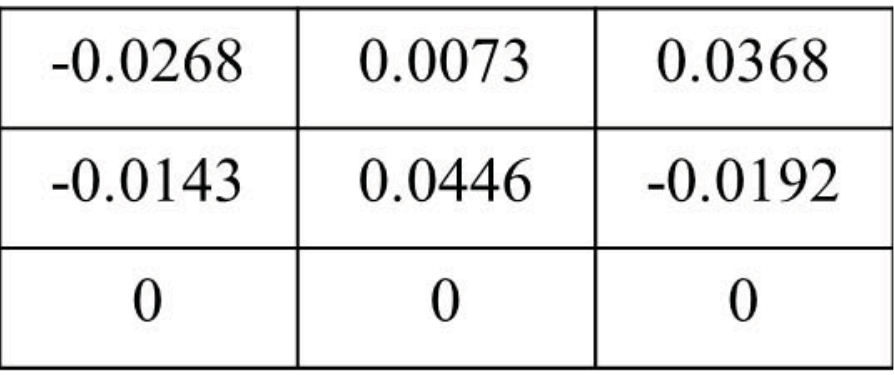}
\label{fig_second_case}}
\caption{(a)-(k) are the visualization of classification results of different methods on Th1. (l) shows the projection learned by our method on Th1.}
\label{fig:3}
\end{figure*}

\begin{table*}[t!]
\caption{Classification accuracies (\%) of different methods on the three-ring data. (Note: NC is the abbreviation of the nearest neighbor classify \cite{bishop2006pattern}.}
\scriptsize
\centering
\label{tab:1}
\begin{tabular}{|c|c|c|c|c|c|c|c|c|c|c|c|c|c|c|c|c|}
\hline
No. & NC &LRC &CRC &SRC & LRLR & LRRR & SLRR & DLSR & ReLSR & SVM & DRLS & MSRL &CLRS &GReLSR & ALPR \\
\hline
\hline
Th1& 93.13 &33.00 &33.33 &39.40  & 36.13 &35.67 &36.13 &63.60  &63.40 &99.20 &77.33 & 66.60 &36.07 &34.06  & \textbf{99.93}\\
Th2& 38.33 &41.46 &33.33 &42.27  & 31.00 &30.93 &31.00 &41.26  &68.13 &67.80 &37.93 & 69.27 &32.00 &33.53  & \textbf{99.87}\\
\hline
\end{tabular}
\end{table*}

\subsection{Experiments and analysis on real-world databases}
In this subsection, five benchmark databases listed in Table II are chosen to evaluate the effectiveness of our method.
\begin{table}[t!]
\caption{Descriptions of the used real-world databases.}
\label{tab:1}
\centering
\footnotesize
\begin{tabular}{|p{1.5cm}|p{3cm}|p{1cm}|p{1.2cm}|}
\hline
Database       & \# Sample per class & \# Class & \# Feature\\
\hline
\hline

COIL100        & 72       & 100      &1024   \\
PIE            & 164-170    &68      &1024     \\
LFW            & 11-20    & 86       &1024  \\
Scene\_SPM       & 210-410    &15       &3000   \\
CIFAR-10       &6000      &10        &1000\\
\hline
\end{tabular}
\end{table}

(1) \textbf{COIL100 object database}{\footnote{COIL100 database is available at: \url{http://www.cs.columbia.edu/CAVE/software/softlib/coil-100.php}}} \cite{nayar1996columbia}: COIL100 database is one of the most popular benchmarks for object classification. It is composed of 7200 images provided by 100 objects, in which each object has 72 images with different poses. Fig.5 (a) shows some example images of the database. Each image used in the experiments was normalized and resized to $32\times32$ with black background in advance. For this database, we randomly select 10, 15, 20, and 25 samples of each class to form the training set and treat the remaining samples as the test set, respectively.

(2) \textbf{CMU pose, illumination, and expression (PIE) face database} \cite{sim2002cmu}: PIE database is one of the challenging databases for face recognition since it contains over 40, 000 images with various poses, illumination conditions, and expressions, etc. In our experiments, we compare different methods on a subset of PIE which totally contains 11554 images of 68 individuals. Each class has nearly 170 samples with 5 different poses. Fig.5 (b) shows some typical images of the database. For computational efficiency, each image was pre-resized to $32\times32$ and then stacked into a vector with 1024 dimensions. For this database, we randomly select 10, 15, 20, and 25 samples per class as the training set and treat the remaining samples as the test set for experiments, respectively.

(3) \textbf{Labeled Faces in the Wild (LFW) face database} \cite{learned2016labeled}: The LFW database is more challenge than the PIE face database since all images are directly collected from the web with different poses, backgrounds, expressions, illuminations, and image acquirement devices, etc. In our experiments, a subset which contains 1251 cropped face images provided by 86 persons is chosen for comparison \cite{wang2012sparse}. There are about 11-20 samples in each class. Some typical images from the same class are shown in Fig.5 (c). Similarly, each image was transformed into a $32\times32$ matrix and then stacked into a vector. Then 5, 6, 7, and 8 samples are randomly chosen from each class to form the training set and the reaming samples are regarded as the test set accordingly.

(4) \textbf{Fifteen Scene Categories (Scene15) database}{\footnote{The Fifteen Scene Categories database is available at: \url{http://www-cvr.ai.uiuc.edu/ponce_grp/data/}}} \cite{lazebnik2006beyond}: The Scene15 database is widely chosen to evaluate different methods for the scene classification task. The 4485 images are naturally collected from 15 common scenes in daily life, such as street, office, store, highway, living room and kitchen, etc. For each scene, there are about 210-410 natural samples. Fig. 5 (d) shows some typical images of the database. It is not suitable to directly evaluate different methods on the original images because they have many differences in size, intensity, shape, and background, etc. In our experiments, we compare different methods on its feature-level database by following the experimental settings in \cite{zhang2017discriminative}, in which all samples are represented by their spatial pyramid features with 3000 dimensions. We refer to the feature-level database as Scene\_SPM for convenience. Similarly, 10, 20, 30, and 40 samples of each class are randomly selected to form the training set and the remaining samples are regarded as the test set, respectively.

(5) CIFAR-10 database \cite{krizhevsky2009learning}: The CIFAR-10 database is a popular large-scale image database, which consists of 50000 training images and 10000 test images from 10 classes. The size of the original color images in the database is $32\times 32$. Some typical images are shown in Fig.5 (e). In our work, we first exploit \emph{k}-means based feature extraction method{\footnote{The feature extraction code of \emph{k}-means is available at:
\url{http://ai.stanford.edu/~acoates/papers/kmeans_demo.tgz}}} \cite{coates2011analysis} to extract the features of CIFAR-10 database, and then utilize the principal component analysis (PCA) \cite{jolliffe2011principal} algorithm to reduce the feature of each sample to 1000 dimensions to improve the computational efficiency. We refer to the extracted features of CIFAR-10 as K-means-CIFAR10. On this dataset, several well-known deep convolutional network based classification methods, including ResNet with 110 layers \cite{he2016deep}, simple fast convolutional (SFC) \cite{macedo2018simple}, deep linear discriminant analysis (DeepLDA) \cite{dorfer2015deep}, and DensetNet \cite{huang2017densely}, are also compared.
\begin{table*}[!htb]
\caption{Mean classification accuracies (\%) of different methods on the COIL100 database. Note: (1) bold numbers denote the best results; (2) we directly list the results of MSRL reported in \cite{zhang2017marginal}.}
\label{tab:4}
\centering
\footnotesize
\begin{tabular}{|c|c|c|c|c|c|c|c|c|c|c|c|c|c|c|c|}
\hline
No. &LRC &CRC &SRC & SVM & LRLR & LRRR & SLRR & DLSR & ReLSR & DRLS & MSRL &CLRS &GReLSR & ALPR \\
\hline
\hline
10	&82.77	&77.80	&84.30	&83.99	&55.78	&65.86	&58.81	&82.47	&81.66	&77.52	&\textbf{88.40}	&79.29    &78.99    &87.69\\
15	&88.82	&82.31	&85.07	&89.04	&61.04	&69.61	&64.99	&87.55	&86.17	&81.44	&\textbf{93.32}	&83.35    &83.50    &91.72\\
20	&91.82	&84.89	&87.86	&92.12	&67.26	&72.08	&69.04	&92.57	&89.11	&88.15	&\textbf{95.87}	&85.85    &86.23    &94.37\\
25	&93.64	&86.61	&90.67	&93.89	&72.22	&74.73	&73.01	&93.28	&93.23	&90.06	&\textbf{97.15}	&87.96    &88.28    &95.97\\
\hline
\end{tabular}
\end{table*}
\begin{table*}[!htb]
\caption{Mean classification accuracies (\%) of different methods on the PIE database. Note: (1) bold numbers denote the best results; (2) we directly list the results of MSRL reported in \cite{zhang2017marginal}.}
\label{tab:4}
\footnotesize
\centering
\begin{tabular}{|c|c|c|c|c|c|c|c|c|c|c|c|c|c|c|c|}
\hline
No. &LRC &CRC &SRC & SVM & LRLR & LRRR & SLRR & DLSR & ReLSR & DRLS & MSRL &CLRS &GReLSR & ALPR \\
\hline
\hline
10	&75.16	&86.33	&72.48	&77.87	&73.06	&86.67	&86.88	&82.55	&87.53	&84.70	&89.51	&89.58    &87.22    &\textbf{91.14}\\
15	&84.60	&90.86	&82.62	&86.44	&80.26	&89.99	&90.25	&89.34	&91.89	&89.40	&93.39	&92.93    &91.43    &\textbf{94.49}\\
20	&89.62	&92.98	&85.66	&92.65	&82.27	&91.96	&92.57	&92.28	&93.89	&92.32	&95.02	&94.50    &93.48    &\textbf{96.00}\\
25	&91.87	&93.94	&89.90	&93.74	&88.22	&93.45	&93.88	&94.16	&95.19	&93.82	&95.96	&95.39    &94.71    &\textbf{96.67}\\
\hline
\end{tabular}
\end{table*}
\begin{table*}[!htb]
\caption{Mean classification accuracies (\%) of different methods on the LFW database. Note: bold numbers denote the best results.}
\label{tab:4}
\footnotesize
\centering
\begin{tabular}{|c|c|c|c|c|c|c|c|c|c|c|c|c|c|c|c|}
\hline
No. &LRC &CRC &SRC & SVM & LRLR & LRRR & SLRR & DLSR & ReLSR & DRLS & MSRL &CLRS &GReLSR & ALPR \\
\hline
\hline
5	&29.73	&30.12	&29.38	&26.04	&30.24	&33.37	&30.57	&27.90	&31.81	&26.26	&32.34	&36.91   &37.31    &\textbf{37.39}\\
6	&32.18	&31.44	&32.51	&29.52	&33.29	&35.24	&34.15	&30.80	&34.45	&28.07	&35.68	&40.48   &40.10    &\textbf{41.39}\\
7	&34.53	&32.51	&33.64	&30.60	&34.96	&35.59	&34.36	&33.73	&37.70	&33.97	&38.45	&41.91   &42.72    &\textbf{43.27}\\
8	&37.23	&34.55	&35.12	&33.14	&35.59	&36.52	&35.64	&36.80	&40.37	&34.52	&42.58	&44.39   &44.55    &\textbf{45.93}\\
\hline
\end{tabular}
\end{table*}
\begin{table*}[htb]
\caption{Mean classification accuracies (\%) of different methods on the Scene\_SPM database. Note: bold numbers denote the best results.}
\label{tab:4}
\footnotesize
\centering
\begin{tabular}{|c|c|c|c|c|c|c|c|c|c|c|c|c|c|c|c|}
\hline
No. &LRC &CRC &SRC & SVM & LRLR & LRRR & SLRR & DLSR & ReLSR & DRLS & MSRL &CLRS &GReLSR & ALPR \\
\hline
\hline
10	&87.75	&87.64	&87.60	&85.09	&81.08	&86.02	&84.44	&87.77	&88.04	&86.98	&88.86	&89.41   &89.84    &\textbf{90.86} \\
20	&92.21	&92.02  &91.99	&91.30	&89.49	&88.24	&89.53	&91.49	&92.04	&93.53  &93.60	&94.06   &94.21    &\textbf{95.25} \\
30	&93.64	&94.02	&92.89	&92.90	&86.59	&87.72	&89.75	&93.50	&93.36	&94.70	&95.44	&95.75   &95.83    &\textbf{96.66} \\
40	&94.97	&94.64	&95.49	&93.43	&91.38	&90.34	&91.07	&94.22	&95.79	&95.21	&96.52	&96.71   &96.90    &\textbf{97.62} \\
\hline
\end{tabular}
\end{table*}
\begin{table}[htb]
\caption{Classification accuracies (ACC) (\%) of different methods on the K-means-CIFAR10 database. For the four deep learning based methods, we directly list their reported results.}
\label{tab:4}
\footnotesize
\centering
\begin{tabular}{|c|c|c|c|c|c|}
\hline
Method & ACC  &Method & ACC &Method & ACC \\
\hline
\hline
LRC	   &58.87    &DLSR	 &67.15	  &GReLSR	   &70.49\\
CRC	   &56.35    &ReLSR  &64.82   &ALPR        &72.37\\
SRC	   &54.67    &SVM	 &71.41	  &ResNet      &93.57\\
LRLR   &65.14	 &DRLS	 &66.95   &SFC         &92.19\\
LRRR   &65.21    &MSRL   &70.83   &DeepLDA     &92.71\\
SLRR   &65.14    &CLRS   &70.12   &DensetNet   &94.81\\
\hline
\end{tabular}
\end{table}

For the first four databases, we repeatedly perform different methods 20 times and report their mean classification accuracies for fair comparison. For the CIFAR-10 database, we implement all methods on the same 50000 training samples and 10000 test samples. The experimental results of different methods on the above five databases are enumerated in Table III-Table VII. From the experimental results, we can find that our method obtains much better performance than the other methods in most cases. In addition, the following interesting points can be obtained according to the experimental results:

(1) We can find that DLSR, ReLSR, CLRS, and GReLSR generally outperform LRLR, LRRR, and SLRR on the above five databases, which proves the effectiveness of the $\varepsilon $-dragging technique and discriminative target learning technique. In other words, learning a more flexible target matrix with large margins of different classes is beneficial to learn a more discriminative projection for classification.

(2) From these four tables, it is obvious that MSRL and the proposed method always perform much better than ReLSR. As analyzed in the previous section, MSRL and the proposed method are the two extensions of ReLSR, which exploit the local geometric information of data to guide the projection learning. Therefore, the experimental results prove that preserving the local geometric structure of data during the linear regression is also significant and enables the two methods to learn a more discriminative projection.

(3) The proposed method and MSRL obtain comparative good performance on the PIE and Scene-SPM database. While on the LFW database, the proposed method significantly outperforms MSRL. From Fig.5 (c), we can find that images of the same class also have very large differences in the LFW database. Thus it is difficult to capture the intrinsic geometric structure of data especially using the unsupervised approach. In other words, MSRL cannot find the intrinsic nearest neighbor relationships to guide the projection learning, and thus cannot guarantee the satisfactory performance. Compared with MSRL, our method overcomes this issue by exploiting a supervised approach to adaptively capture the intrinsic similarity relationships among samples of the same class, which plays a positive guiding role in the projection learning. Meanwhile, as analyzed in the previous section, the proposed method has the potential to adaptively select those important features from data for feature extraction and effectively reduce the negative influence of noise, which is also beneficial to improve the classification performance. These two effective approaches encourage the proposed method to obtain a better performance than MSRL on the LFW database.

(4) From Table VII, we can obviously find that all the deep convolutional network based methods achieve much better performance than the conventional methods. This demonstrates that the deep convolutional network based methods can extract more discriminative features than the exploited unsupervised feature extraction method, \emph{i.e}, \emph{k}-means. Among all of the conventional methods, the proposed method still outperforms all the other methods, which also proves that the proposed method can learn a more discriminative projection than the other conventional methods.
\begin{figure}[!htb]
\centering
\subfloat[COIL100]{\includegraphics[width=3.5in]{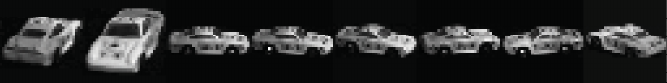}
\label{fig_first_case}}\vspace{-.4cm}
\hfil
\subfloat[PIE]{\includegraphics[width=3.5in]{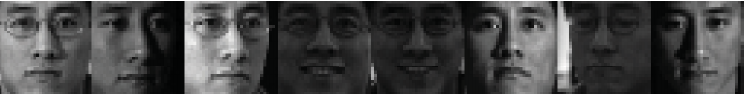}
\label{fig_second_case}}\vspace{-.4cm}
\hfil
\subfloat[LFW]{\includegraphics[width=3.5in]{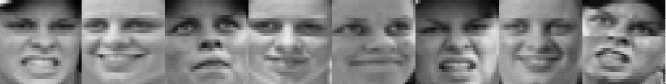}
\label{fig_second_case}}\vspace{-.4cm}
\hfil
\subfloat[Scene15]{\includegraphics[width=3.5in]{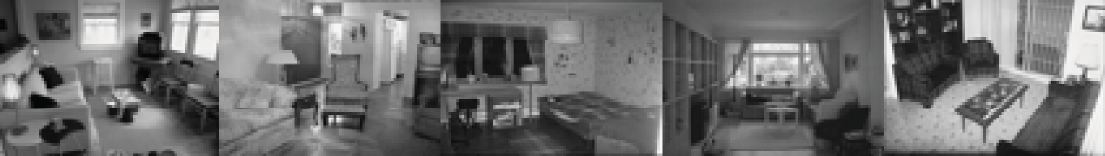}
\label{fig_second_case}}\vspace{-.4cm}
\hfil
\subfloat[CIFAR-10]{\includegraphics[width=3.5in]{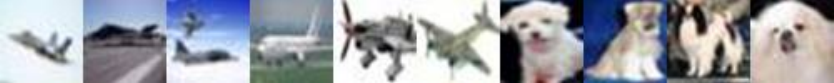}
\label{fig_second_case}}\vspace{-.1cm}
\caption{Typical images of the used real-world databases.}
\label{fig:3}
\end{figure}

\subsection{Parameter sensitivity and selection}
Generally, for some methods, selecting the optimal penalty parameters is crucial to achieve satisfactory performance on different databases. In this section, we mainly analyze the sensitivity of parameter selection of the proposed method and then provide a simple strategy to select the optimal parameters. From (8), we can find that the proposed method only contains two penalty parameters, \emph{i.e.}, ${\lambda _1}$ and ${\lambda _2}$, which are regularized on the nearest neighbor preserving term and the feature selection term, respectively. To analyze the sensitivity of the classification performance to these two parameters, firstly, a large candidate range $\{ {{{10}^{ - 5}},{{10}^{ - 4}},{{10}^{ - 3}},{{10}^{ - 2}},{{10}^{ - 1}},1,{{10}^1},{{10}^2},{{10}^3},{{10}^4},{{10}^5}}\}$ is defined for the two penalty parameters. Secondly, we conduct several experiments to show the relationships of the classification accuracies (\%) and different values of the two parameters on the first four databases. Fig.6 shows the classification accuracies versus the two parameters. From these figures, it is obvious that when parameter ${\lambda _1}$ is selected from the range of $\left[ {{{10}^{ - 5}},1} \right]$, and parameter ${\lambda _2}$ locates in the proper range, such as $\left[ {0.1,1} \right]$ on the COIL100 database, $\left[{0.1,1} \right]$ on the PIE database, $\left[ {0.1,1} \right]$ on the LFW database, and $\left[ {{{10}^{ - 5}},1} \right]$ on the Scene\_SPM database, respectively, the proposed method can obtain almost constant classification accuracy. This demonstrates that the proposed method is insensitive to the selection of ${\lambda _1}$ to some extent.

As far as we know, it is still an open problem to adaptively select the optimal parameters for different databases. In this paper, we exploit a simple approach based on the grid search to find the two optimal parameters \cite{wen2018inter,wen2018robust}. According to the previous analysis, we first define the candidate range $\left[ {{{10}^{ - 5}},1} \right]$ for the two parameters. Then we fix parameter ${\lambda _1}$ as 0.1 since the proposed method is insensitive to it, and perform the proposed method with different values of ${\lambda _2}$ selected from the coarse candidate range. In this way, we can find the latent optimal ${\lambda _2}$ from the candidate range. Then we fix ${\lambda _2}$ with the obtained latent optimal value and perform the proposed again to find the optimal value of ${\lambda _1}$ from the same candidate range. Finally, we can obtain the best combination of the two parameters for experiments.
\begin{figure*}[htb]
\centering
\subfloat[COIL100]{\includegraphics[width=1.5in]{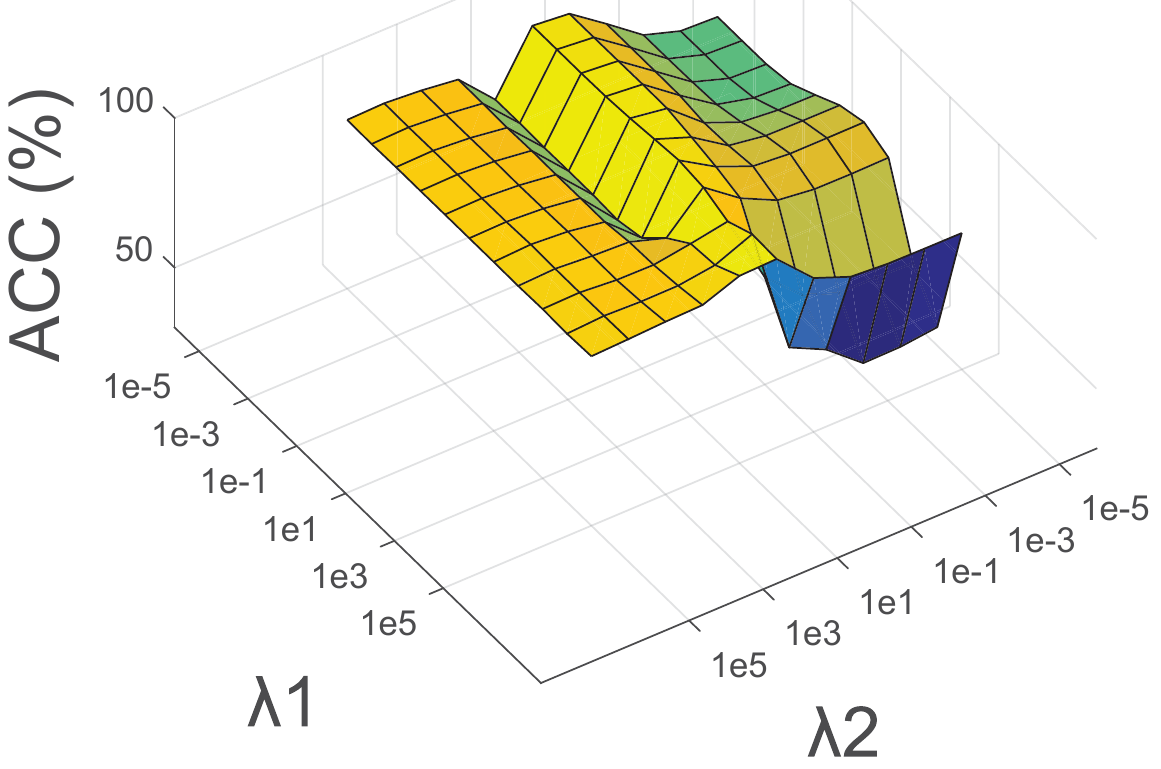}
\label{fig_first_case}}
\hfil
\subfloat[PIE]{\includegraphics[width=1.5in]{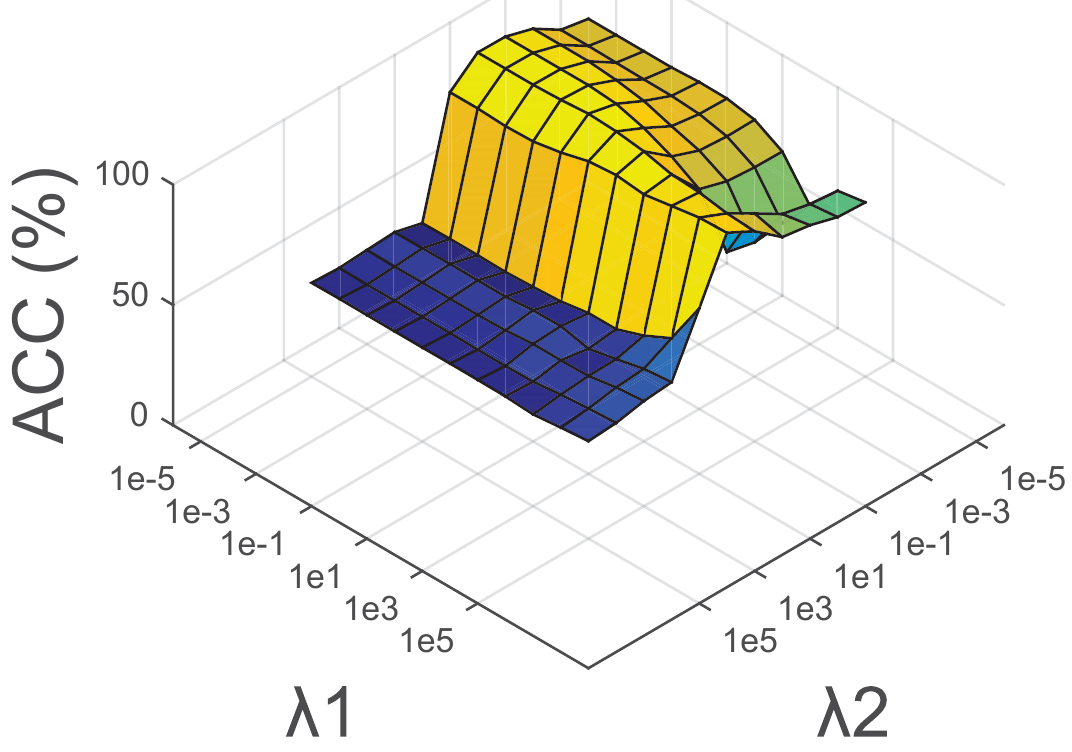}
\label{fig_second_case}}
\hfil
\subfloat[LFW]{\includegraphics[width=1.5in]{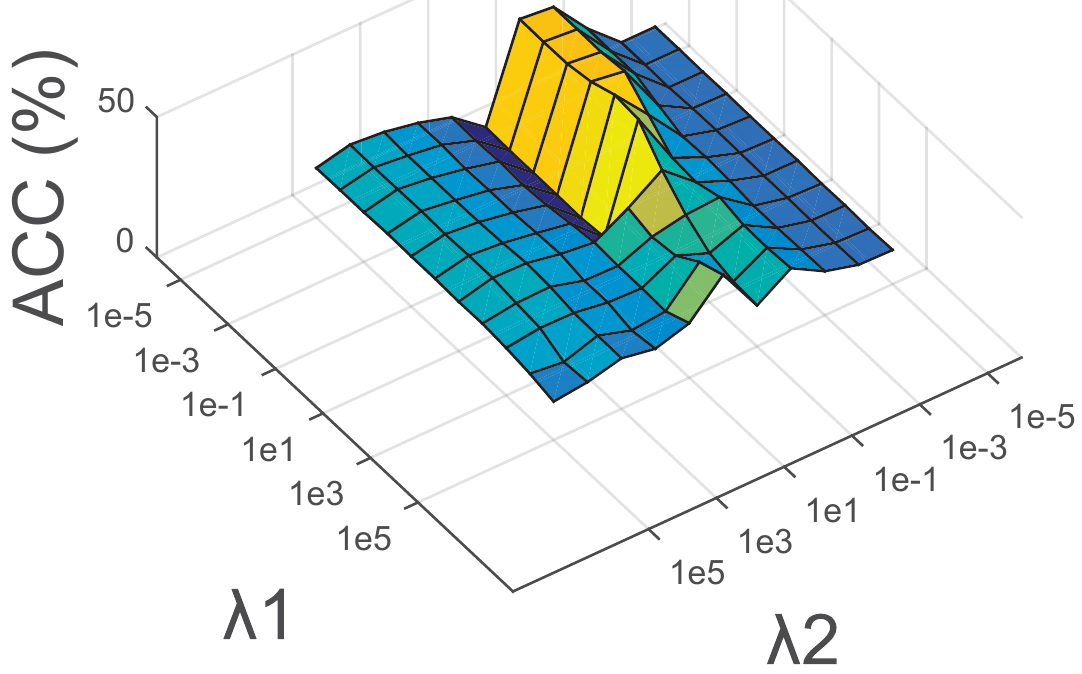}
\label{fig_second_case}}
\hfil
\subfloat[Scene\_SPM]{\includegraphics[width=1.5in]{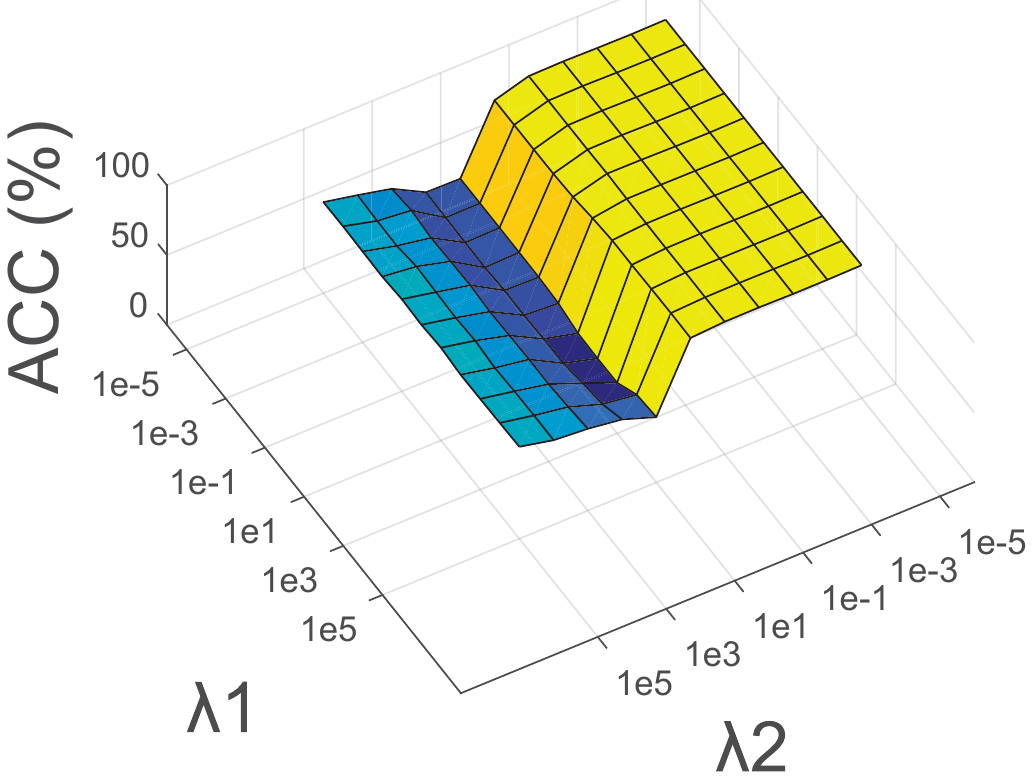}
\label{fig_second_case}}
\caption{Classification accuracy (ACC) (\%) versus ${\lambda _1}$ and ${\lambda _2}$ on the (a) COIL100, (b) PIE, (c) LFW, and (d) Scene\_SPM databases, in which 20, 20, 7, and 20 samples are randomly selected from each class to form the training set, respectively.}
\label{fig:3}\vspace{-.8cm}
\end{figure*}
\begin{figure*}[htb]
\centering
\subfloat[COIL100]{\includegraphics[width=1.5in]{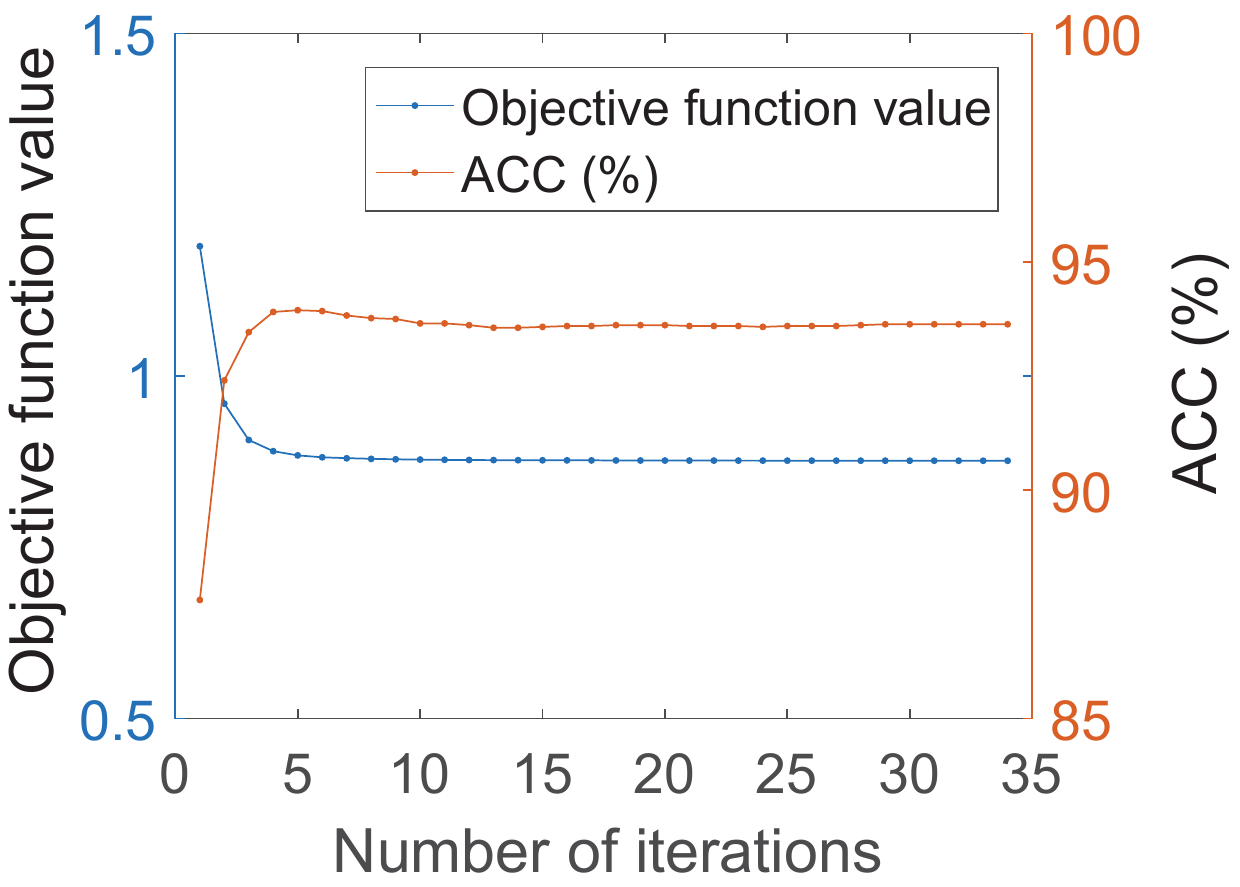}
\label{fig_first_case}}
\hfil
\subfloat[PIE]{\includegraphics[width=1.5in]{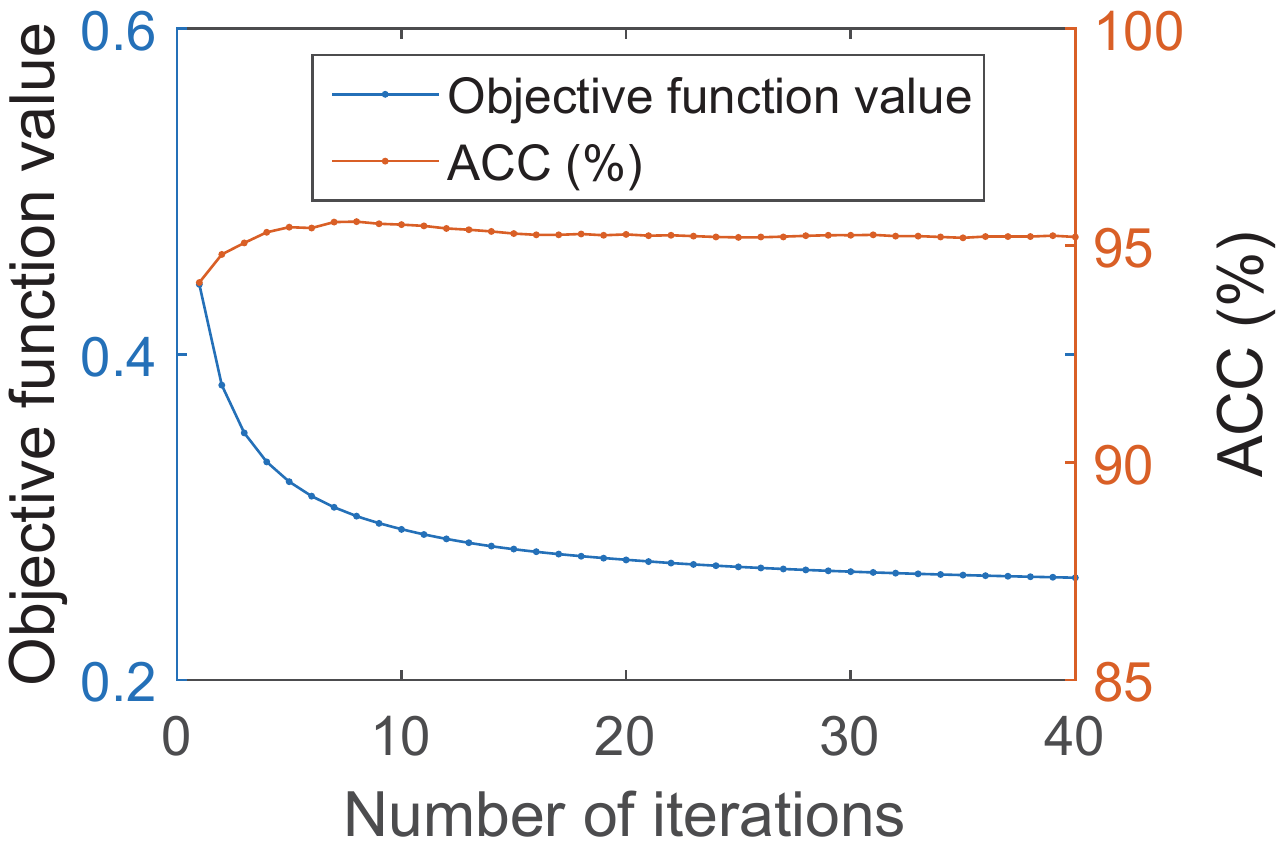}
\label{fig_second_case}}
\hfil
\subfloat[LFW]{\includegraphics[width=1.5in]{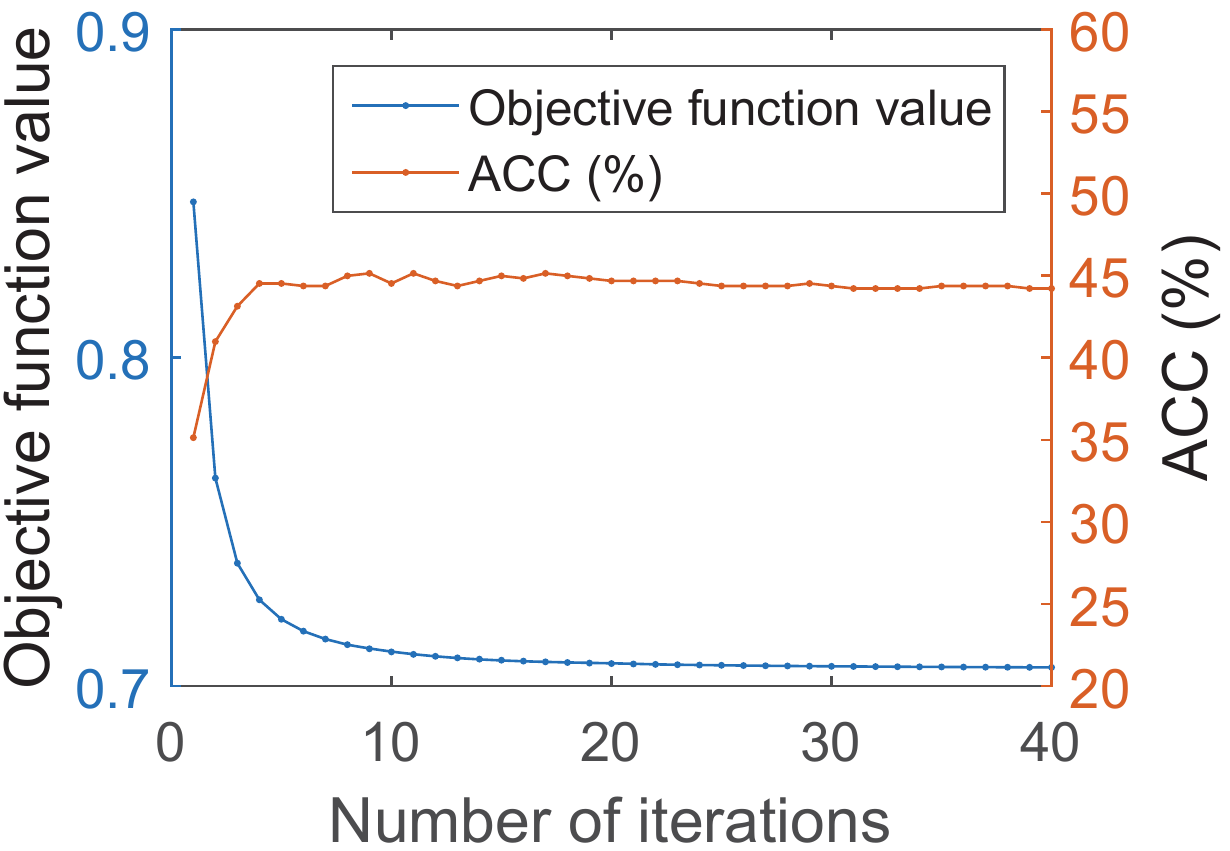}
\label{fig_second_case}}
\hfil
\subfloat[Scene\_SPM]{\includegraphics[width=1.5in]{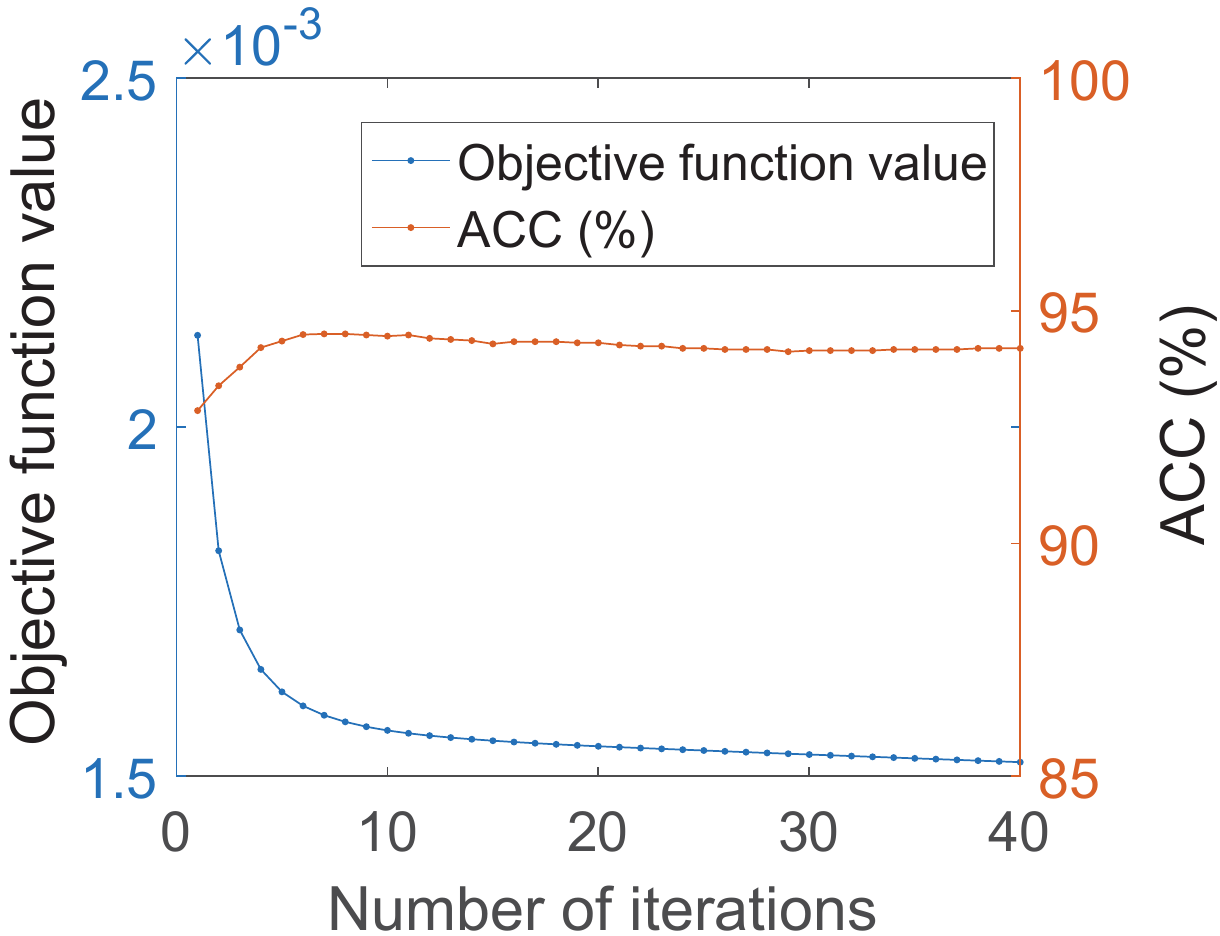}
\label{fig_second_case}}
\caption{Objective function value and classification accuracy (ACC) (\%) versus the iteration step on the (a) COIL100, (b) PIE, (c) LFW, and (d) Scene\_SPM databases, in which 20, 20, 7, and 20 samples are randomly selected from each class to form the training set, respectively.}
\label{fig:3}
\end{figure*}
\subsection{Experiments of convergence study}
In this section, we mainly conduct some experiments to further prove the convergence property of the proposed optimization approach in Algorithm 1. In Fig. 7, we have plotted the objective function values and classification accuracies versus the iteration steps on the COIL100, PIE, LFW, and Scene\_SPM databases, respectively, in which the objective function value is directly calculated as $obj = {{\left( {\left\| {T - {X^T}W} \right\|_F^2 + {\lambda _1}Tr\left( {{W^T}{S_W}W} \right) + {\lambda _2}{{\left\| W \right\|}_{2,1}}} \right)} \mathord{\left/
 {\vphantom {{\left( {\left\| {T - {X^T}W} \right\|_F^2 + {\lambda _1}Tr\left( {{W^T}{S_W}W} \right) + {\lambda _2}{{\left\| W \right\|}_{2,1}}} \right)} {\left\| X \right\|_F^2}}} \right.
 \kern-\nulldelimiterspace} {\left\| X \right\|_F^2}}$ according to the objective function (8). From these figures, it is obvious that the objective function value is monotonically decreasing till to the stationary point with the iteration increasing, which proves the point of \textbf{Theorem 3}. Meanwhile, we can also find that the classification accuracy increases obviously until the objective function value converges to the stationary point, which demonstrates that the proposed method can finally find the local optimal solution when the objective function converges.

\section{Conclusion}
We proposed an effective linear regression method for classification in this paper. The proposed method improves the discriminability of projection through three approaches. Firstly, we adaptively learn a more flexible target matrix with large margins between the correct and incorrect classes for regression. Secondly, we replace the conventional `Frobenius norm' with the sparse $l_{2,1}$ norm to constrain the projection, which enables the proposed method to select the most important features from the original high-dimensional data for feature extraction. Thirdly, we introduce a novel supervised graph regularization term to guide the projection learning. Compared with the conventional unsupervised graph learning approach, the supervised approach presented in our paper is more effective in preserving the intrinsic nearest neighbor relationships of each class. Most importantly, the discriminative target learning, intrinsic graph learning, and projection learning are neatly integrated into one joint learning framework, which enables the method to obtain the global optimal projection for classification so as to obtain a better performance. The effectiveness of the proposed method has been sufficiently proved on the synthetic database and many real-world databases.

\bibliographystyle{IEEEtran}
\bibliography{main}

\end{document}